\documentclass[final,12pt]{clear2026} 

\usepackage{natbib}
\usepackage{amssymb}
\usepackage{xcolor}
\usepackage{colortbl}

\usepackage{hyperref}

\usepackage{booktabs}
\usepackage{bm}
\hypersetup{colorlinks=true,linkcolor=blue,filecolor=blue,citecolor=blue}

\usepackage{tikz}
\usepackage{relsize}
\usepackage{etoolbox}

\newtheorem{assumption}{A\hspace{-0.35em}}
\newtheorem{intassumption}{Assumption}
\numberwithin{intassumption}{assumption}

\usepackage{algorithm}
\usepackage{algorithmic}

\newcommand{\E}{\mathbb{E}}
\newcommand{\PP}{\mathbb{P}}
\newcommand{\bX}{\boldsymbol X}
\newcommand{\bx}{\boldsymbol x}
\newcommand{\indep}{\perp \!\!\! \perp}

\newcommand{\val}{\text{val}}

\newcommand{\sumn}{\frac{1}{n}\sum_{i=1}^n}
\newcommand{\II}{\mathbb{I}}

\DeclareMathOperator*{\argmin}{arg\,min}

\title[DML for Counterfactual Conformal Prediction Under Runtime Confounding]{Debiased Machine Learning for Conformal Prediction of Counterfactual Outcomes Under Runtime Confounding}
\usepackage{times}

\clearauthor{%
 \Name{Keith Barnatchez} \Email{keithbarnatchez@g.harvard.edu}\\
 \addr Department of Biostatistics, Harvard T.H. Chan School of Public Health
 \AND
 \Name{Kevin P. Josey} \Email{kevin.josey@cuanschutz.edu}\\
 \addr Department of Biostatistics and Informatics, Colorado School of Public Health
 \AND
  \Name{Rachel C. Nethery} \Email{rnethery@hsph.harvard.edu}\\
 \addr Department of Biostatistics, Harvard T.H. Chan School of Public Health
 \AND 
 \Name{Giovanni Parmigiani} \Email{gp@jimmy.harvard.edu}\\
 \addr Department of Data Science, Dana Farber Cancer Institute
}

\begin{document}

\maketitle

\begin{abstract}%
Data-driven decision making frequently relies on predicting counterfactual outcomes.  In practice, researchers commonly train counterfactual prediction models on a source dataset to inform decisions on a possibly separate target population. Conformal prediction has arisen as a popular method for producing assumption-lean prediction intervals for counterfactual outcomes that would arise under different treatment decisions in the target population of interest. However, existing methods require that every confounding factor of the treatment-outcome relationship  used for training on the source data is additionally measured  in the target population, risking miscoverage if important confounders are unmeasured in the target population. In this paper, we introduce a computationally efficient debiased machine learning framework that allows for valid prediction intervals when only a subset of confounders is measured in the target population, a common challenge referred to as runtime confounding. Grounded in  semiparametric efficiency theory, we show the resulting prediction intervals achieve  desired coverage rates with faster convergence compared to standard methods. Through numerous synthetic and semi-synthetic experiments, we demonstrate the utility of our proposed method.
\end{abstract}


\section{Introduction}
\label{introduction}


Data-driven decision-support tools (DSTs) are experiencing rapid growth across diverse domains, including personalized medicine, marketing, and social services \citep{musen2021clinical,chouldechova2018case,fischer2024bridging}. Of particular value are DSTs  which predict individual-level \textit{counterfactual outcomes} arising from different possible actions, or treatments, performed by the decision-maker. Recognizing that decision-makers often require \textit{uncertainty quantification} around individual-level counterfactual predictions, an emerging interdisciplinary literature has developed at the intersection of causal inference, statistics and machine learning focused on constructing robust prediction intervals for counterfactual outcomes predicted from flexible, nonparametric models. 

Beginning with the seminal work of \cite{lei2021conformal}, there has been a growing effort to extend tools from conformal prediction to enable the formation of prediction intervals of counterfactual outcomes and individual treatment effects \citep{yang2024doubly,liu2024multi,alaa2023conformal,gao2025bridging,schroder2025conformal}. Despite the large interest in methods for constructing robust prediction intervals in settings ranging from covariate shift to surrogate outcomes, numerous practical challenges remain unaddressed. One particular challenge is \textit{runtime confounding},  where only a subset of the covariates needed to adjust for confounding of the treatment-outcome relationship are measured on units in the target population where counterfactual predictions are desired \citep{coston2020counterfactual}.  

Runtime confounding arises frequently in practice, and is typically induced when one is able to collect extensive covariate information for training in a source population, but collecting  information on this full set of covariates in the target population of interest is cost prohibitive or infeasible. For instance, in personalized medicine, electronic health records may contain detailed patient histories during model training, but point-of-care decisions often rely on limited covariate measurements \citep{deschepper2025literature,collins2023prediction}. Similarly, in marketing applications, customer profiles built from historical data may be unavailable due to privacy regulations when making real-time recommendations \citep{bleier2020consumer}. Naively restricting models to train on only the set of confounders available in both the source and target populations risks yielding inaccurate prediction intervals when  discarded variables serve as confounders of the treatment-outcome relationship. Despite this, existing methods typically assume full access to confounders in the source and target populations, leaving researchers with little recourse to address this challenge.

\textbf{Contributions.} In this work, we propose methods for performing conformal prediction  of counterfactual outcomes which address the critical challenge of runtime confounding. Our proposed approach leverages tools from semiparametric theory and debiased machine learning (DML) \citep{park2025semiparametric}, ensuring constructed intervals attain valid coverage under a modest set of conditions   relative to competing methods. We provide a computationally efficient implementation of our approach, which avoids challenges commonly faced by DML methods. Through numerous numerical experiments, we compare our proposed method to alternative approaches based on existing popular frameworks, demonstrating conditions under which our method tends to outperform the latter methods. We additionally derive valid loss functions for counterfactual quantile regression under runtime confounding settings, where the resulting predictions can be used to construct quantile conformity scores \citep{romano2019conformalized} within our proposed framework. Although not our primary contribution, we additionally provide a weighted conformal prediction method capable of addressing runtime confounding.

\textbf{Related work.} Our work is situated at the intersection of causal inference, conformal prediction and transfer learning.
Leveraging results from \cite{tibshirani2019conformal} and \cite{romano2019conformalized},  \cite{lei2021conformal} introduced weighted quantile conformal prediction to construct intervals for counterfactuals, addressing  covariate shift across treatment levels.  Subsequent work has extended these ideas, implementing doubly-robust methods addressing covariate shift and multi-study settings \citep{yang2024doubly,liu2024multi},
accounting for surrogate outcomes \citep{gao2025bridging}, and basing scores on meta-learners of counterfactual outcomes and treatment \citep{alaa2023conformal}. Our work  explicitly addresses covariate shift across treatment levels within the source population,  as well as between target and source populations, while allowing for incomplete confounder information in the target population.

The causal transfer learning literature aims to address distribution shifts between source and target population data in order to estimate marginal and conditional causal effects \citep{shyr2025multi,bica2022transfer,colnet2024causal,rojas2018invariant,voter2025counterfactual}.  Recent work has developed doubly-robust methods for unknown shifts based in semiparametric theory \citep{graham2024towards,zeng2025efficient},  with numerous extensions accommodating multi-source data and privacy constraints \citep{han2025federated,han2023multiply} and the incorporation of surrogate outcomes 
\citep{kallus2025role}.  We contribute to this literature by deriving valid loss functions for causal quantile regression that enable learning target population counterfactual quantile functions, accounting for covariate shift across treatment levels and populations.

The problem of runtime confounding for point prediction of counterfactual outcomes was formalized by \cite{coston2020counterfactual}.  Our work is the first to extend this problem to the construction of \textit{prediction intervals} through semiparametric efficient  conformal prediction. By allowing for covariate shift across the target and source populations we extend the work of \cite{coston2020counterfactual}, who only considered covariate shift across treatment levels within the source population. More broadly, our setting is connected to a wider literature on counterfactual prediction in which the deployed prediction rule may condition on only a subset of confounders, including work motivated by transportability and time-varying covariate information \citep{boyer2025estimating,keogh2024prediction}.

\section{Problem Setting and Background}
\label{sec:prob-setting}

We consider a setting where researchers are interested in the relationship between a categorical treatment variable $A$ taking on values in a set $\mathcal{A}$ and an outcome of interest $Y$. Complete information on $Y$ and $A$ is provided for all units in a source population dataset, while both $Y$ and $A$  are unavailable in a target population dataset, consistent with a setting where target population members have not yet received treatment. Source population membership is denoted by the  indicator variable $S$. It is assumed there is a set of baseline covariates $\bX =(\bm V, \bm U)$ that are fully measured in the source population dataset, whereas only a subset of these covariates, $\bm V$, are measured in the target population. 
The induced data structure can be characterized by the observational unit,
\[
\bm O_i = (S_iY_i, \ S_iA_i, \ S_i\bm U_i, \ \bm V_i, \ S_i) \sim \PP, \ \ \ i = 1,\ldots,n,
\]
where we adopt the convention that for any random variable $Z$, its observed value is $S Z + (1-S) \texttt{NA}$ to make explicit that $Y, A$ and $\bm U$ are only observed when the source data indicator $S=1$. Following \cite{coston2020counterfactual}, we refer to this setting as \textbf{runtime confounding}, since $\bm U$ may contain potential confounding factors of the relationship between $A$ and $Y$.

Table \ref{tab:data-structure} further summarizes this data structure. We note that such a data structure could arise in both single- or multi-source settings. For instance, the above data structure could arise from a setting where an initial batch of data on $Y, A$ and $\bX$ is collected from a single site, and additional observations on only $\bm V$ are collected by the same site, possibly according to a different sampling strategy that induces covariate shift. Similarly, the above structure could arise if the target population observations $\bm V$ are collected at an external site which lacks the capacity to measure the additional confounding variables $\bm U$, but wishes to use models trained in the source population to construct prediction intervals for its units.
\begin{table}[h!]
    \centering
    \resizebox{0.3\linewidth}{!}{
    \begin{tabular}{ccccc}
    \toprule
        $Y$ & $A$ & $\bm V$ & $\bm U$ & $S$  \\
        \midrule
        \rowcolor{blue!25} $Y_1$ & $A_1$ & $\bm V_1$ & $\bm U_1$ & 1 
         \\
        \rowcolor{blue!25} \vdots & \vdots & \vdots & \vdots & \vdots 
        \\ 
        \rowcolor{blue!25} $Y_{n_1}$ & $A_{n_1}$ & $\bm V_{n_1}$ & $\bm U_{n_1}$ & 1  
        \\
        \rowcolor{red!25} \texttt{NA} & \texttt{NA} & $\bm V_{n_1+1}$ & \texttt{NA} & 0 
        \\
        \rowcolor{red!25} \vdots & \vdots & \vdots & \vdots & \vdots 
        \\
        \rowcolor{red!25} \texttt{NA} & \texttt{NA} & $\bm V_{n_1+n_0}$ & \texttt{NA} & 0  
                \\
         \bottomrule
    \end{tabular}}
        \caption{Observed data structure.}
    \label{tab:data-structure}
\end{table}
\\[0.4em]
\textbf{Objective:} We fix interest on the construction of prediction intervals for counterfactual outcomes of subjects in the target population. Following the Rubin potential outcomes framework \citep{little2000causal}, we let $Y_i(a)$ denote the counterfactual outcome that subject $i$ would experience under treatment level $A_i=a$. Our primary goal is to form prediction intervals $C_a(\bm V)$ that cover counterfactual outcomes in the target population with a desired coverage probability:
\[
\PP(Y(a) \in C_a(\bm V)|S=0) \geq 1-\alpha, \ \ a \in \mathcal{A},
\]
where $C_a(\bm V)$ is a function of $\bm V$ since $\bm U$ is unavailable for observations in the target populations. Naively ignoring $\bm U$ throughout training in the source population can yield intervals with severe miscoverage when $\bm U$ includes important confounders between $Y$ and $A$. In turn, our methods leverage $\bm U$ in the source population, while allowing for $\bm U$ to be unmeasured in the target population.

Multiple prior works have  considered settings where $\mathcal{A}=\{0,1\}$ and interest lies in constructing intervals for individual treatment effects (ITEs) $Y(1)-Y(0)$ \citep{lei2021conformal,yang2024doubly,alaa2023conformal}.
We focus on constructing intervals for counterfactual outcomes over ITEs for numerous reasons. While ITEs are useful estimands in many decision-support settings, in many  settings ITEs are less informative for decision-makers. For instance, when costs are associated with different treatment levels, one may prefer a less costly treatment  so long as their predicted outcome is predicted to exceed a lower bound, regardless of whether the more expensive treatment would yield a greater response. Further, in settings with many treatment levels, the utility of numerous pairwise ITE intervals is less apparent. 
In Appendix \ref{ap:ite-methods} we demonstrate how our method can be adjusted to produce intervals for ITEs.

\subsection{Assumptions}

The construction of valid prediction intervals for counterfactual outcomes under runtime confounding requires a set of standard causal inference assumptions, as well as assumptions commonly invoked in causal data fusion problems \citep{degtiar2023review}. We begin with assumptions necessary for forming valid prediction intervals within the source population. 
\begin{assumption}[Positivity]
\label{as:assumption-positivty}
$0<\PP(A=a | \bX=\bx)<1$ for all $\bx$ with positive support
\end{assumption} \vspace{-1.1em}
\begin{assumption}[Consistency] 
\label{as:assumption-consistency} 
$Y = \sum_{a \in \mathcal{A}} \mathbb{I}(A=a)\cdot Y(a)$
\end{assumption} \vspace{-1.1em}
\begin{assumption}[Unconfoundedness]
\label{as:assumption-unconfoundedness}
$Y(a) \indep A | \bX, S=1$
\end{assumption} 

Assumptions \ref{as:assumption-positivty}-\ref{as:assumption-unconfoundedness} are standard assumptions in the causal inference literature \citep{rosenbaum1983central, rubin2005causal}. While the above assumptions are sufficient for the construction of valid prediction intervals in the source data, we require additional assumptions to construct prediction intervals over the target population.

\begin{assumption}[Source exchangeability]
\label{as:source-exch}
$Y(a) \indep S | \bm V$
\end{assumption} \vspace{-1.1em}
\begin{assumption}[Source positivity]
\label{as:source-pos}
$0 < \PP(S=1|\bm V=\bm v) < 1$ for all $\bm v$ with positive support
\end{assumption}

Assumptions \ref{as:source-exch} and \ref{as:source-pos} are standard assumptions in the data fusion 
literature \citep{bareinboim2016causal,degtiar2023review}, where interest typically fixates on using source data to estimate average treatment effects for a separate target population. 
Assumption \ref{as:source-exch} implies all systematic differences in $Y(a)$ across the source and target population are explained by $\bm V$, while
Assumption \ref{as:source-pos} ensures overlap of the distribution of $\bm V$ between the target and source populations. Collectively, the independence Assumptions \ref{as:assumption-unconfoundedness} and \ref{as:source-exch} imply the distribution shift between observational units in the source and target populations arises due to two sources of covariate shift: (i) covariate shift in $\bX$ across treatment levels within the source population, and (ii) covariate shift in $\bm V$ across the source and target populations. 
Note that by naively discarding all of $\bm U$ within the source population and ignoring the possibility of runtime confounding,  researchers would implicitly require a more stringent variant of Assumption \ref{as:assumption-unconfoundedness} that instead requires $Y(a)\indep A | \bm V,S=1$. Our set of Assumptions relaxes this requirement, allowing for the possibility that $\bm U$ is a confounder of $Y$ and $A$ so long as all systematic differences in counterfactual outcomes $Y(a)$ across the target and source populations are explained by the always-observed covariates $\bm V$.

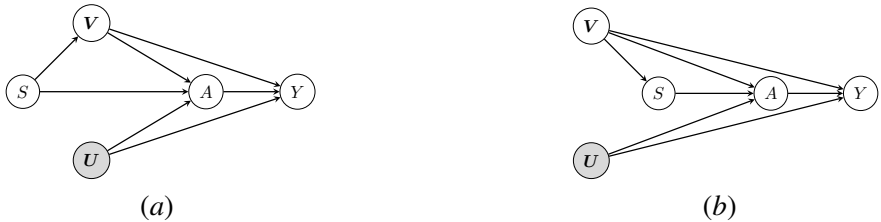
\begin{figure}[h!]
\centering
\subfigure[]{
    \resizebox{0.27\linewidth}{!}{%
    \begin{tikzpicture}[node distance=0.8cm, every node/.style={transform shape}]
        \node[circle,draw,minimum size=0.7cm] (V) at (-0.5,1.5) {$\bm V$};
        \node[circle,draw,minimum size=0.7cm,fill=gray!30] (U) at (-0.5,-1.5) {$\bm U$};
        \node[circle,draw,minimum size=0.7cm] (S) at (-2,0){$S$};
        \node[circle,draw,minimum size=0.7cm] (A) at (2,0) {$A$};
        \node[circle,draw,minimum size=0.7cm] (Y) at (4,0) {$Y$};
        \draw[->, thick, >=stealth] (V) -- (A);
        \draw[->, thick, >=stealth] (S) -- (V);
        \draw[->, thick, >=stealth] (V) -- (Y);
        \draw[->, thick, >=stealth] (U) -- (A);
        \draw[->, thick, >=stealth] (U) -- (Y);
        \draw[->, thick, >=stealth] (S) -- (A);
        \draw[->, thick, >=stealth] (A) -- (Y);
    \end{tikzpicture}}}
    \hspace{8em}
\subfigure[]{
    \resizebox{0.27\linewidth}{!}{
        \begin{tikzpicture}[node distance=0.8cm, every node/.style={transform shape}]
        \node[circle,draw,minimum size=0.7cm] (V) at (-2,1.5) {$\bm V$};
        \node[circle,draw,minimum size=0.7cm,fill=gray!30] (U) at (-2,-1.5) {$\bm U$};
        \node[circle,draw,minimum size=0.7cm] (S) at (-0.5,0){$S$};
        \node[circle,draw,minimum size=0.7cm] (A) at (2,0) {$A$};
        \node[circle,draw,minimum size=0.7cm] (Y) at (4,0) {$Y$};
        \draw[->, thick, >=stealth] (V) -- (A);
        \draw[->, thick, >=stealth] (V) -- (S);
        \draw[->, thick, >=stealth] (V) -- (Y);
        \draw[->, thick, >=stealth] (U) -- (Y);
        \draw[->, thick, >=stealth] (U) -- (A);
        \draw[->, thick, >=stealth] (S) -- (A);
        \draw[->, thick, >=stealth] (A) -- (Y);
    \end{tikzpicture}}}
    \caption{Two possible directed acyclic graphs consistent with Assumptions \ref{as:assumption-unconfoundedness} and \ref{as:source-exch}. Runtime confounding is induced when $\bm U$ is unobserved when $S=0$. }
    \label{fig:dag_grid}
    \vspace{-2em}
\end{figure} 

We note that in settings where data are collected by a single site---with an initial batch $(S=1)$ that includes measurements on $\bX$ used for training, and a second batch $(S=0)$ only containing measurements on $\bm V$---Assumption \ref{as:source-exch} will often be plausible, since batch membership will commonly arise from data collection logistics, rather than factors that systematically influence the outcome of interest. This batch collection setting is exclusively considered in \cite{coston2020counterfactual},  who implicitly invoke Assumption \ref{as:source-exch}. 
We instead adopt a more general framing that treats the source and target datasets as arising from distinct sites, so as to cover broader data collection regimes. We note that the single-site batch setting considered by \citep{coston2020counterfactual} can be viewed as a special case of this formulation, where Assumption \ref{as:source-exch} makes explicit what sources of distribution shift are permitted across batches. 
Like Assumption \ref{as:assumption-unconfoundedness}, Assumption \ref{as:source-exch} should be informed by subject-matter expertise in these broader settings, and we discuss avenues for sensitivity analyses in Section \ref{sec:discussion}.
Further, in Appendix \ref{sec:app-ind-discussion} we show that Assumption \ref{as:source-exch} is \textit{implied} by (i) a weaker conditional independence $Y(a) \indep S | \bm{X}$, and (ii) a covariate shift condition on $\bm U$ across study populations $\bm{U} \indep S | \bm{V}$. We additionally discuss examples of settings in which we expect Assumption \ref{as:source-exch} to hold or be violated. Figure \ref{fig:dag_grid} presents example causal directed acyclic graphs consistent with Assumptions \ref{as:assumption-unconfoundedness} and \ref{as:source-exch}.

\vspace{-1.1em}
\section{Conformal Prediction}
\label{sec:conformal-background}

Conformal prediction \citep{vovk2005algorithmic} is a general statistical framework enabling the construction of valid prediction intervals under minimal distributional assumptions. 
The framework's generality
has spurred a growing field of research into implementations that account for common challenges arising in prediction tasks.  We provide a brief overview of the framework here, and recommend  \cite{fontana2023conformal} and \cite{angelopoulos2024theoretical} for extensive reviews of the many active research areas in this field.

Briefly considering a single-source prediction problem with covariates $\bX$ and outcome $Y$ for simplicity, an object central to conformal prediction is the \textit{conformity score} $R(Y,\bX)$. At a high-level, $R(Y,\bX)$ is defined so that extreme values imply a lack of agreement, or conformity, between the actual outcome $Y$ and predicted value based on $\bX$. Conversely,  smaller magnitude values imply higher conformity. Numerous conformity scores are used in practice, though two  popular choices are the absolute residual
$R(Y,\bX) = |Y - \hat \E(Y|\bX)|$ and the quantile regression score $R(Y,\bX) = \max \{Y - \hat Q_{\alpha/2}(\bX), \hat Q_{1-\alpha/2}(\bX) - Y\}$ \citep{romano2019conformalized}, where $Q_\alpha(\bX)$ is the $\alpha$ quantile of the conditional distribution $Y|\bX$.

For a fixed choice of conformity score, let $r_{1-\alpha}$ denote its theoretical $1-\alpha$ quantile. Conformal prediction is driven by the observation that for intervals of the form $\hat C(\bX) = \{y: R(y,\bX) \leq r_{1-\alpha}\}$, by construction $\PP(Y \in \hat C(\bX)) = 1-\alpha$. In turn, conformal prediction methods fixate interest on estimation of the unknown quantity $r_{1-\alpha}$, with many methods performing adjustments to improve finite-sample performance or, in certain circumstances, provide distribution-free finite sample guarantees that ensure a coverage lower bound of $1-\alpha$ \citep{angelopoulos2023conformal}. Such finite-sample guarantees are generally unattainable when there is covariate shift between the source and target dataset whose form must be estimated \citep{barber2023conformal}. To reconcile the impossibility of finite-sample guarantees, our proposed methods leverage tools from semiparametric theory to enhance their finite sample behavior.  
\subsection{Constructing Conformity Scores Under Runtime Confounding}

While our proposed methods are valid for any choice of conformity score, we briefly present recommendations for constructing conformity scores under runtime confounding. In such settings, conformity scores take the form $R_a(Y(a), \bm V)$, reflecting the partial availability of covariates in the target population and fact that scores will be indexed by counterfactual outcomes occurring under different treatment levels $a \in \mathcal{A}$. Notice $R_a(Y(a),\bm V)$ will only be observable for source units with treatment level $A=a$, for whom the consistency Assumption \ref{as:assumption-consistency} implies $R_a(Y(a),\bm V)=R_a(Y,\bm V)$. In turn, we interchangeably use both notations as appropriate.  For brevity, we consider analogues of the absolute residual and quantile conformity scores.

Under Assumptions \ref{as:assumption-positivty}-\ref{as:source-pos}, \cite{coston2020counterfactual}, who focus on counterfactual prediction of $Y(a)$ under runtime confounding, leverage the observation that $\E[Y(a)|\bm V,S=0] = \E[\E(Y|A=a,\bX,S=1)|\bm V,S=1]$. This observation implies one can construct point predictors of $Y(a)$ by (i) estimating $\mu_a(\bm X) = \E(Y|A=a,\bX,S=1)$ within the source population, and (ii) further regressing $\hat \mu_a(\bX)$ on $\bm V$ within the source population, obtaining predictions $\hat \eta_a(\bm V)$. Such a procedure implies the conformity score $R_a(Y(a),\bm V) = |Y(a) - \hat \eta_a(\bm V)|$, computable for source population units receiving treatment level $A=a$. \cite{coston2020counterfactual} additionally propose a doubly-robust procedure that enables faster convergence, but they do not focus on the construction of prediction intervals. In either case, our proposed methods in Section \ref{sec:methods} provide a valid procedure to quantify the uncertainty around the resulting predictions.

We next propose a method for constructing quantile conformity scores, which relies on estimation of the  quantile function of the conditional distribution $Y(a) | \bm V, S=0$.
Let $Q_{a,\alpha}(\bX)$ and $Q_{a,\alpha}(\bm V)$ denote the $1-\alpha$ quantiles of $Y(a)|\bm X,S=0$ and $Y(a)|\bm V,S=0$, respectively. Estimation of $Q_{a,\alpha}(\bm V)$ requires additional care, since $\E[Q_{a,\alpha}(\bX) | \bm V,S=1] \neq Q_{a,\alpha}(\bm V)$ due to the nonlinearity of quantile functions. The result below provides a valid loss function for estimation of $Q_{a,\alpha}(\bm V)$ in the presence of runtime confounding. 
\begin{proposition}
\label{prop:wgtd-quant}
Suppose $Q_{a,\alpha}(\bm v)$ satisfies $\PP(Y(a) \leq Q_{a,\alpha}(\bm v) | \bm V=\bm v,S=0)=1-\alpha$ for all $\bm v$ with positive support. Then, under Assumptions \ref{as:assumption-positivty}-\ref{as:source-pos}, $Q_{a,\alpha}(\bm V)$ additionally satisfies
\begin{align}
\hspace{-0.5em}Q_{a,\alpha}(\bm V) = \argmin_{\tilde Q_{a,\alpha}} \E\left[w_a(\bm O)  \rho_{\alpha}(Y - \tilde Q_{a,\alpha}(\bm V)) \right], \label{eq:quant-prop}
\end{align}
where $\rho_{\alpha}(x) = \alpha 
|x| \mathbb{I}(x\geq 0) + (1-\alpha)|x|\mathbb{I}(x<0)$ is the pinball loss function and 
\[
w_a(\bm O) := \frac{\mathbb{I}(A=a)S(1-\kappa(\bm V))}{g_a(\bX)\kappa(\bm V)}.
\]
\end{proposition}
where $g_a(\bX) := \PP(A=a|\bX,S=1)$ and $\kappa(\bm V) := \PP(S=1|\bm V)$. Proposition \ref{prop:wgtd-quant} suggests one can consistently estimate the conditional quantile function $Q_{a,\alpha}(\bm V)$ through minimization of the empirical analogue of the weighted pinball loss function \citep{koenker1978regression} appearing in Proposition \ref{prop:wgtd-quant}. Notably, \eqref{eq:quant-prop} represents a weighted quantile regression among source units with treatment level $A=a$, enabling the use of existing software packages which support weights or simple augmentations to existing estimation procedures. 
Intuitively, the weight $w_a(\bm O)$ can be interpreted as a product of two distinct adjustment factors accounting for two sources of covariate shift. The first, $1/g_a(\bX)$, is an inverse probability of treatment weight that adjusts for covariate shift in $\bX$ across treatment levels within the source population. The second, $(1-\kappa(\bm V))/\kappa(\bm V)$, is an inverse odds weight that re-weights the source population to resemble the target population with respect to $\bm V$, accounting for covariate shift in $\bm V$ across these two populations. 
Relative to quantile scores based on unweighted quantile regression, we expect intervals based on scores formed from our proposed weighted quantile loss function to exhibit improved finite-sample performance, since the weights $w_a(\bm O)$ enable consistent estimation of $Q_{a,\alpha}(\bm V)$.

\section{Methods}
\label{sec:methods}

Given a fixed conformity score $R_a(Y, \bm V)$, we aim to construct intervals of the form $\hat C_a(\bm V) = \{
y: R_a(y,\bm V) \leq \hat r_{a,\alpha}\}$, where $\hat r_{a,\alpha}$ is an estimate of the $1-\alpha$ quantile of scores $R_a$ in the target distribution which satisfies
\begin{equation}
    \label{eq:target-quantile}
    \PP(R_a(Y(a), \bm V) \leq \hat r_{a,\alpha} | S=0) = 1-\alpha.
\end{equation}
(\ref{eq:target-quantile}) implies that  construction of valid prediction intervals for $Y(a)$ in the target population requires accurate estimation of $r_{a,\alpha}$. In this Section, we present a roadmap for constructing efficient estimators of $r_{a,\alpha}$.

\subsection{Identification}

Although conformity scores cannot be directly observed within the target population,
Assumptions \ref{as:assumption-positivty}-\ref{as:source-pos} ensure $r_{a,\alpha}$ can be expressed in terms of scores formed in the source population. We present two novel identification functionals which enable estimation of $r_{a,\alpha}$ in Theorem \ref{thm:identification} below.

\begin{theorem}
\label{thm:identification}
Let $R_a(Y,\bm V)$ be a generic conformity score for  $Y(a)$, and suppose $r_{a,\alpha}$ satisfies
$$
\PP(R_a(Y(a),\bm V) \leq r_{a,\alpha} | S=0) = 1-\alpha.
$$
Under Assumptions \ref{as:assumption-positivty}-\ref{as:source-pos}, $r_{a,\alpha}$ additionally satisfies
\begin{align}
&   \E[m_a(r_{a,\alpha},\bm V)  \ | \ S=0] = 1-\alpha \label{eq:thm1-gcomp} \\
&   \E\left[ w_a(\bm O) \mathbb{I}(R_{a}(Y,\bm V) \leq r_{a,\alpha}) \right] = 1-\alpha. \label{eq:thm1-ipw}
\end{align}
\end{theorem}
where we define $q_a(r,\bX) := \PP(R_a(Y,\bm V) \leq r | \bX, A=a,S=1)$ and $m_a(r,\bm V) := \E[q_a(r,\bX) | \bm V, S=1]$.
Equations (\ref{eq:thm1-gcomp}) and (\ref{eq:thm1-ipw}) are closely related to functionals arising in the transportability literature \citep{zeng2025efficient}, and can be viewed as extensions of identifying functionals from the conformal prediction under covariate shift literature \citep{tibshirani2019conformal,yang2024doubly}. 
Intuitively, the above expressions both address two separate sources of covariate shift: between levels of treatment $A$ among source population units, and between members of the source and target populations. Critically, (\ref{eq:thm1-gcomp}) and (\ref{eq:thm1-ipw}) suggest regression- and weighting-based means for constructing valid prediction intervals of $Y(a)$ in the target population. 

\subsection{Plug-in Estimation}

Given the identification expression (\ref{eq:thm1-gcomp}), a natural approach to constructing prediction intervals for $Y(a)$ is to form a plug-in estimate of $r_{a,\alpha}$ by choosing $\hat r_{a,\alpha}$ to solve the following estimating equation in $r$
    \begin{equation}
    \label{eq:gcomp-est-eq}
        \frac{1}{n_0}\sum_{i:S_i=0} \hat m_a(r,\bm V_i) - (1-\alpha) = 0,  
    \end{equation}
 noting solving \eqref{eq:gcomp-est-eq} requires repeatedly fitting $\hat q_a(r,\bm X)$ and $\hat m_a(r, \bm V)$ with pre-specified learners for potentially many values of $r$. For any $r$, $m_a(r, \bm V)$ can be estimated by first fitting $\hat q_a(r,\bX)$ among units with treatment $A=a$ in the source population, regressing the predicted values on $\bm V$ among source units.
Letting $\PP_n\{f(\bm O)\}:=\sumn f(\bm O_i)$ for generic $f$, one can additionally obtain a weighting-based estimator of $r_{a,\alpha}$ through (\ref{eq:thm1-ipw}) by solving the estimating equation
\begin{align}
\label{eq:weight-est-eq}
& \PP_n\{\hat w_a(\bm O) \II(R_a(Y,\bm V) \leq r_{a,\alpha})\} -  (1-\alpha) = 0,
\end{align}
which circumvents the computational challenge faced by  \eqref{eq:gcomp-est-eq} since $w_a(\bm O)$ does not depend on $r$.
While the above plug-in estimators are consistent for $r_{a,\alpha}$ under correct specification of all relevant nuisance models, the \textit{rate} at which these estimators converge to $r_{a,\alpha}$ will be dictated by the convergence rates of their corresponding nuisance function estimators. These rates can be particularly slow when one chooses to fit all nuisance models with flexible learners, in order to minimize the risk of model misspecification. 
\begin{algorithm*}[t]
\caption{Debiased machine learning split conformal prediction for runtime confounding}
\label{alg:alg-debiased}
\textbf{Input}: Pooled target and source population data $\bm O_i = (Y_i,A_i,\bX_i, S_i)$, desired coverage probability $1-\alpha$, conformity score measure $R_a(Y,\bm V)$
\\[0.8em]
\textbf{Output}: A prediction interval function $\hat C_a(\bm V)$
\\[0.8em]
1. Randomly split the data into a training and calibration set: $\mathcal{D}_\text{train} = \{O_i = (Y_i, A_i, \bm V_i, \bm U_i, S_i),  \ i \in \mathcal{I}_\text{train} \}$, $\mathcal{D}_\text{cal} = \{\bm O_i = (Y_i, A_i, \bm V_i, \bm U_i, S_i),  \ i \in \mathcal{I}_\text{cal} \}$. Further split $\mathcal{D}_\text{train}$ into equally-sized subsets $\mathcal{D}_{\text{train},1}$ and
$\mathcal{D}_{\text{train},2}$
\\[0.8em]
2. Using all of $\mathcal{D}_{\text{train}}$, fit the nuisance functions $\hat g_a$ and $\hat \kappa$ and perform counterfactual prediction of $Y(a)$ to construct conformity scores $R_a(Y,\bm V)$. Using $\mathcal{D}_{\text{train},1}$, obtain an initial estimate of $r_{a,\alpha}$, termed $\hat r_{a,\alpha}^\text{init}$, through a generic estimation algorithm such as weighted conformal prediction \citep{tibshirani2019conformal} 
\\[0.8em]
3. Using $\hat r_{a,\alpha}^\text{init}$ and the second training subset $\mathcal{D}_{\text{train},2}$, obtain estimates $\hat q_a(\hat r_{a,\alpha}^\text{init},\bX)$ and $\hat m_a(\hat r_{a,\alpha}^\text{init},\bm V)$
\\[0.8em]
4. Choose $\hat r_{a,\alpha}$ to solve the estimating equation $\sum_{i \in \mathcal{D}_\text{cal}} \chi_a(r_{a,\alpha}, \bm O_i\  ; \hat \eta_a(\hat r_{a,\alpha}^\text{init} )) =0$ among units in $\mathcal{D}_\text{cal}$
\\[0.8em]
5. Use the resulting estimate $\hat r_{a,\alpha}$ to construct conformal intervals for participants in the target population of the form $\hat C_a(\bm V) = \{y: R_a(y,\bm V) \leq \hat r_{a,\alpha}  \}$
\end{algorithm*}
Following \cite{zeng2025efficient}, \cite{gao2025bridging} and \cite{liu2024multi}, we propose the use of multiply-robust estimators of $r_{a,\alpha}$ that enable faster convergence rates in broader estimation settings.  We turn our attention to the construction of these estimators in the remainder of this Section.

\subsection{Efficiency Theory}

In this section, we present the efficient influence curve (EIC) for $r_{a,\alpha}$. EICs are crucial ingredients for constructing estimators whose convergence rate is dictated by the \textit{product} of nuisance function convergence rates, often allowing for significantly faster estimation of statistical functionals relative to plug-in estimation (\citealt{kennedy2024semiparametric}) while providing partial protection against model misspecification \citep{chernozhukov2018double}. The EIC for $r_{a,\alpha}$ in a fully nonparametric statistical model is presented below, with details on its derivation provided in  Appendix \ref{sec:proofs}.
\begin{theorem}
\label{thm:eif-thm}
Let $\eta_a(r) := (q_a(r),m_a(r),g_a,\kappa)$ and suppose $\bm O \sim \PP$. Under Assumptions \ref{as:assumption-positivty}-\ref{as:source-pos}, the efficient influence curve for $r_{a,\alpha}$ in a  nonparametric model for the observed data distribution $\PP$ is proportional to
\begin{align}
& \chi_a(r_{a,\alpha},  \bm O\  ; \eta_a(r_{a,\alpha})) 
 := \nonumber \\   
& (1-S)(m_a(r_{a,\alpha},\bm V) - (1-\alpha)) 
+
     \frac{S(1-\kappa(\bm V))}{\kappa(\bm V)} 
\{q_a(r_{a,\alpha},\bX) - m_a(r_{a,\alpha},\bm V)\} \nonumber \\
& 
+  w_a(\bm O) \{\II(R_a(Y,\bm V)\leq r_{a,\alpha}) - q_a(r_{a,\alpha},\bX) \}, \label{eq:eic}
\end{align}
where $\E[\chi_a(r_{a,\alpha}, \bm O\  ; \eta_a(r_{a,\alpha}))]=0$.
\end{theorem}
Since the true EIC is mean-zero, we omit the proportionality constant when presenting $\chi_a$ since we will ultimately leverage this moment condition to construct estimators for $r_{a,\alpha}$.

%

\subsection{Debiased Machine Learning Estimation}

The efficient influence curve from  Theorem \ref{thm:eif-thm} can be leveraged to construct efficient estimators of $r_{a,\alpha}$. We follow the framework of \textit{debiased machine learning}, where one chooses $\hat r_{a,\alpha}$ to solve an estimating equation implied by the moment condition in Theorem \ref{thm:eif-thm} \citep{kennedy2024semiparametric}.

The moment condition $\E[\chi_a(r_{a,\alpha},\bm O ; \eta_a(r_{a,\alpha}))]=0$  from Theorem \ref{thm:eif-thm} suggests one can obtain a valid estimate  of $r_{a,\alpha}$ by choosing $\hat r_{a,\alpha}$ to solve $\PP_n\{\chi_a(\hat r_{a,\alpha}, \bm O\  ; \hat \eta_a(\hat r_{a,\alpha}))\} =0.$
However, naively solving the estimating equation in this manner would require iteratively estimating the nuisance functions $q_a$ and $m_a$ for potentially many values  of $r$. To avoid the computational costs associated with this approach, we follow \cite{gao2025bridging} and construct debiased estimators for $r_{a,\alpha}$ based on the DML framework from \cite{kallus2024localized} which allows for $q_a$ and $m_a$ to be estimated at a single, initial estimate of $r_{a,\alpha}$, drastically reducing the computational costs that would be required from repeated estimation of these nuisance functions. This localized construction introduces a preliminary estimator $\hat r^{\mathrm{init}}_{a,\alpha}$, for which $n^{-1/4}$-consistency suffices \citep{kallus2024localized}. By contrast, a non-localized implementation would directly re-estimate $q_a(r,\cdot)$ and $m_a(r,\cdot)$ across candidate values of $r$, so no separate initial estimator is needed. Our proposed split conformal prediction approach, which yields an efficient estimator $\hat r_{a,\alpha}$ and corresponding prediction interval $\hat C_a(\bm V)$  is outlined in Algorithm \ref{alg:alg-debiased}.  

While Algorithm 1 employs split conformal prediction for computational tractability, our framework extends to full conformal prediction by solving $\PP_n \chi_a(r,\bm O; \hat \eta_a)$ without data splitting, as in \cite{yang2024doubly}. This avoids efficiency losses from sample splitting but requires Donsker-type regularity conditions on the nuisance function classes along with the aforementioned increased computational cost.
\subsection{Coverage Properties}
Given a means to construct prediction intervals, we turn our attention to the asymptotic coverage properties of our proposed methods. Since the formation of prediction intervals in our proposed setting requires estimation of the quantity $r_{a,\alpha}$ in \eqref{eq:target-quantile}, one would expect accurate estimation of $r_{a,\alpha}$ should ensure valid coverage of $Y(a)$. Theorem \ref{thm:dr-thm} provides a formal characterization of this notion.
\begin{theorem}
\label{thm:dr-thm}
Suppose that $\hat \kappa(\bm V)$, $\hat g_a(\bm X)$ and $\PP(S=1|\bX)$ are all bounded within $(\varepsilon,1-\varepsilon)$ for some $\varepsilon>0$.
If  $\hat C_a(\bm V)$ is constructed according to Algorithm \ref{alg:alg-debiased}, then
\begin{align*}
\PP(Y(a) \in \hat C_a(\bm V) | S=0) = 1-\alpha + O_\PP(1/\sqrt n \ + \ R_n),    \text{ where}
\end{align*}
\begin{align*}
    R_n &= \sup_{r}|| \hat q_a(r,\cdot) - q_a(r,\cdot)|| \cdot ||\hat g_a - g_a || + \sup_r|| \hat m_a(r,\cdot) - m_a(r,\cdot) || \cdot ||\hat \kappa - \kappa ||.
\end{align*}
\end{theorem}
The structure of the remainder term $R_n$ implies the coverage error shrinks quickly as long as either the estimated outcome-related models ($q_a, m_a$) or the propensity score models ($g_a, \kappa$) converge  sufficiently fast. For example, if all four nuisance functions are estimated using flexible learners with modest convergence rates of $n^{-1/4}$, the product of errors will be of order $n^{-1/2}$. This implies the coverage gap shrinks at the parametric $n^{-1/2}$ rate---the fastest possible rate when $\eta_a(r)$ must be estimated \citep{kennedy2024semiparametric} and a dramatic improvement over plug-in estimators, whose convergence would be limited to the slower $n^{-1/4}$ rate. Due to the protections against misspecification and slow convergence rates, estimators arising from the DML framework are frequently referred to as \textit{doubly-} or \textit{multiply-robust} \citep{kennedy2024semiparametric}.
Beyond the rate conditions on $R_n$, Theorem 4 implies a model-multiple robustness property: consistency of $\hat{r}_{a,\alpha}$ is guaranteed whenever at least one nuisance function in each of the pairs $(m_a, \kappa)$ and $(q_a, g_a)$ is consistently estimated, irrespective of the convergence behavior of the remaining components.
Such improvements are crucial in runtime confounding settings, where numerous nuisance functions need to be modeled.
\vspace{-0.8em}
\section{Experiments}
\label{sec:simulation}

\subsection{Simulated Data}
\label{sec:num-exp}

To assess the performance of our proposed methods, we conducted a simulation study extending the setup considered in \citet{coston2020counterfactual}, who focused on efficient point prediction of $Y(a)$ under the same runtime confounding setting we study. We generate data according to the runtime confounding data structure implied by Figure \ref{fig:dag_grid} and Table \ref{tab:data-structure}, letting $\mathcal{A} = \{0,1\}$ for simplicity and recalling our proposed methods can accommodate categorical $\mathcal{A}$. Additionally, we vary the overall sample size $n$ and number of unmeasured runtime confounders (5, 10, 15), corresponding to cases of mild, moderate and severe runtime confounding. Throughout, we generate $S$ so that $\PP(S=1) = 0.9$, implying for each sample size considered 90\% of observations are from the source population. We extend \cite{coston2020counterfactual} by generating $S$ as a function of $\bm V$ to ensure covariate shift across the source and target populations. 

Full details on the data-generating mechanism, which produces data adhering to the structure in Table \ref{tab:data-structure}, can be found in  Appendix \ref{sec:exp-details}. Replication code can be found at \url{https://github.com/keithbarnatchez/conformal-runtime}. Additional experiments which vary the relative size of the source population are included in Appendix \ref{sec:extra-exp-results}. Across the simulation settings we explore,
we construct 90\% prediction intervals for both  $Y(1)$ and $Y(0)$ in the target data based on both absolute residual and quantile conformity scores. 
We compared our proposed DML procedure to (i) the weighted method implied by equation \eqref{eq:thm1-ipw} and (ii) a DML estimator based on the approach from \cite{yang2024doubly} which ignores runtime confounding by effectively forcing $\bm V = \bm X$.
These two approaches serve as natural comparators, since (i) is the standard approach to addressing distribution shift in conformal prediction problems, and (ii) allows us to investigate the consequences of ignoring runtime confounding while employing an analogous estimation procedure. While not the main contribution of our work, we emphasize that the weighted approach is based on our proposed weights $\hat w_a(\bm O)$, and in turn can be viewed as an additional approach for addressing runtime confounding that we provide.
\begin{figure}
  \centering
  \subfigure[]{\includegraphics[width=0.49\linewidth]{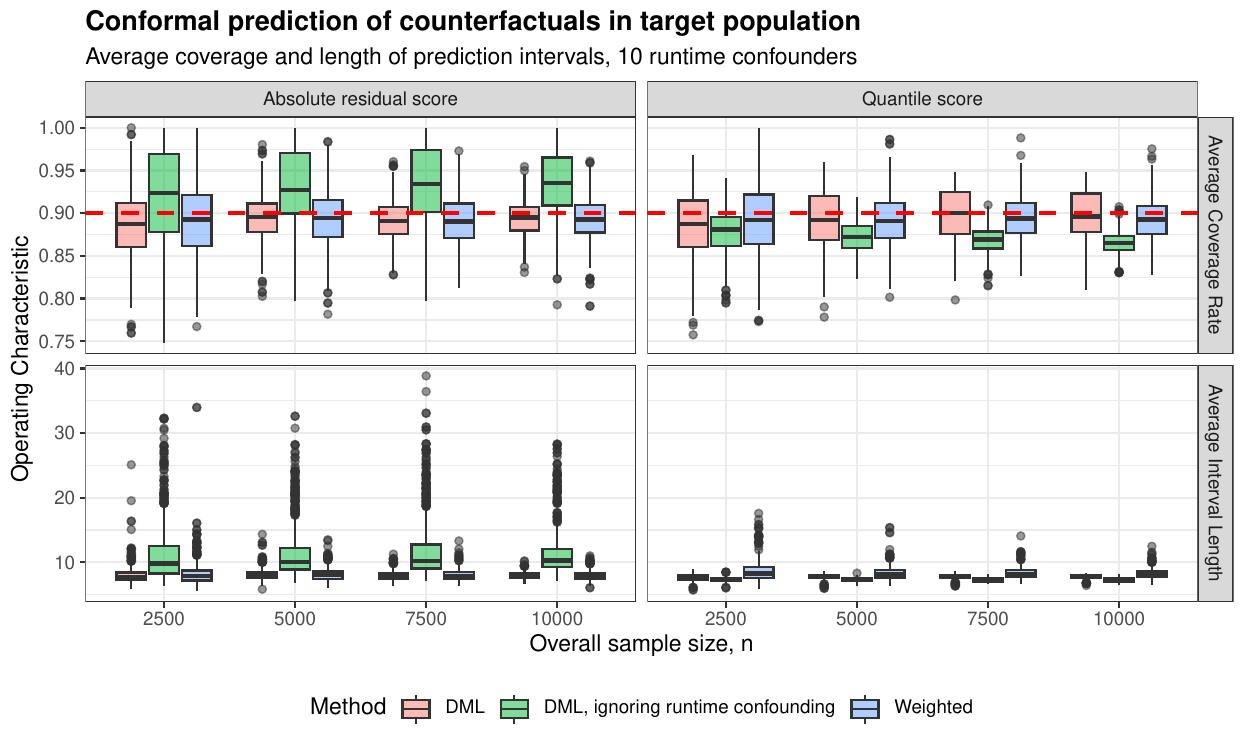}}
  \subfigure[]{\includegraphics[width=0.49\linewidth]{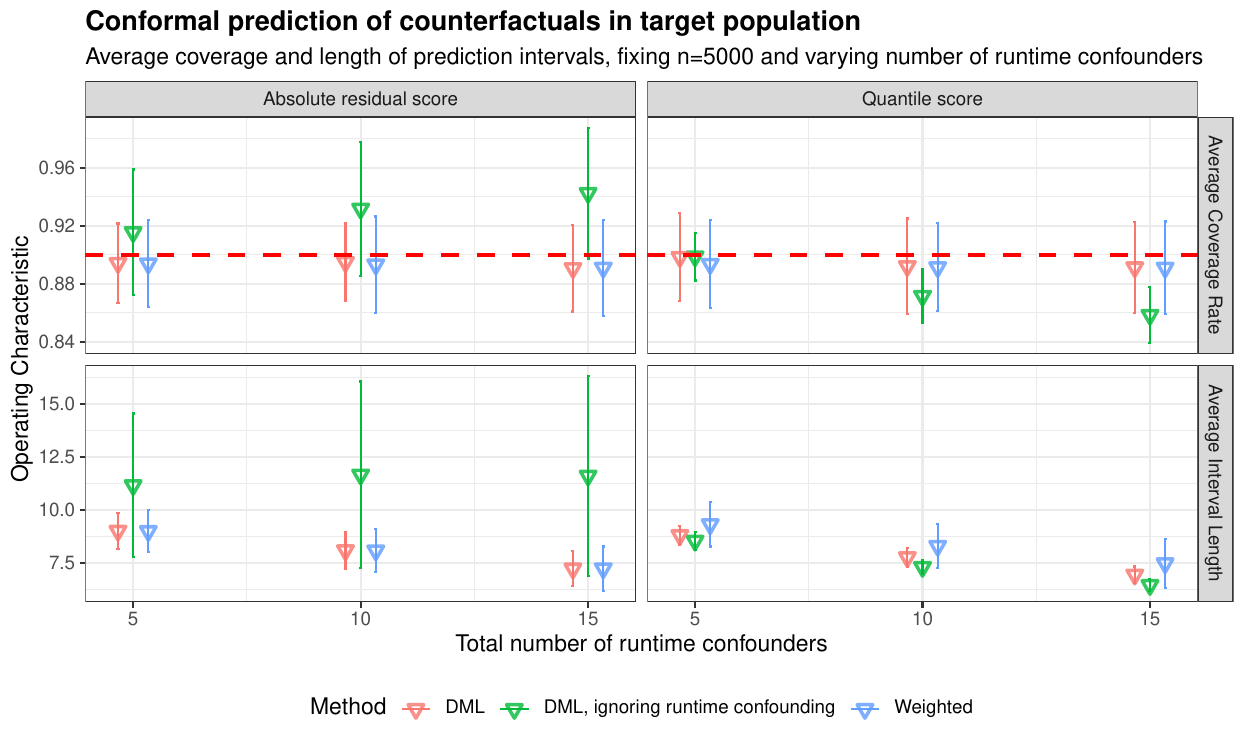}}
  \vspace{-1em}
  \caption{Experiment results, varying $n$ and fixing number of runtime confounders at 10.}
  \label{fig:sim-y1}
  \vspace{-0.5em}
\end{figure}

\textbf{Results:} Figure \ref{fig:sim-y1} displays the results of our experiments. For brevity, we report the empirical coverage rates and interval lengths for $Y(1)$ and $Y(0)$ pooled together, and provide results separately for $Y(1)$ and $Y(0)$ in  Appendix \ref{sec:exp-details}.  We first focus on panel (a), which considers the moderate runtime confounding setting while varying $n$.
We see that  naively ignoring runtime confounding produces intervals which miscover $Y(1)$ and $Y(0)$ at all sample sizes considered. 
Notably, as the overall sample size grows our proposed method rapidly concentrates around the desired 90\% coverage rate, consistent with the results of Theorem \ref{thm:dr-thm}. 
Focusing on panel (b), we see that as runtime confounding becomes more severe, the coverage bias of the DML procedure which ignores runtime confounding worsens, with opposite magnitudes of bias across the two considered conformity scores, highlighting that the coverage bias induced by ignoring runtime confounding can vary systematically and demonstrating the need to adjust for runtime confounding in practice. 
Across all levels of runtime confounding and both conformity scores considered, intervals based on both our DML procedure and weighted conformal with our proposed weights $\hat w_a(\bm O)$ both consistently attain the desired coverage rate of 90\%. Notably, our DML procedure tends to produce intervals that are as or more narrow than the weighted conformal procedure and concentrates rapidly around the  nominal 90\% rate as $n$ grows, highlighting the efficiency of our proposed approach.  
\vspace{-0.7em}
\subsection{Semi-Synthetic Data}
\label{sec:data-appliction}

We examine the performance of our proposed methods on semi-synthetic data from the 2018 Atlantic Causal Inference Conference (ACIC) challenge \citep{carvalho2019assessing}, which is based on the National Study of Learning Minds (NSLM) trial \citep{yeager2019national} and has been used in previous studies of conformal inference for counterfactual outcomes \citep{lei2021conformal}. To ensure access to ground-truth counterfactual outcomes, we generate 1{,}000 synthetic datasets from the ACIC NSLM database following the same approach outlined in \cite{lei2021conformal}, who used this dataset to evaluate weighted conformal prediction methods for individual treatment effects. We enforce runtime confounding by simulating a source population indicator dependent on a subset of covariates in the ACIC NSLM dataset, and assume the analyst only has access to this subset of covariates for the target population. Analogous to Section \ref{sec:num-exp} we fix $\PP(S=1)=0.9$, and repeat this exercise for different overall sample sizes, and vary variables included in $\bm V$ and $\bm U$ to examine increasingly severe cases of runtime confounding that we term mild, moderate and severe. We generate $S$ as a function of $\bm V$ that enforces covariate shift between the target and source populations. Full details on the data generation procedure are provided in  Appendix \ref{sec:exp-details}.
\begin{figure}[t]
    \centering
    \includegraphics[scale=0.45]{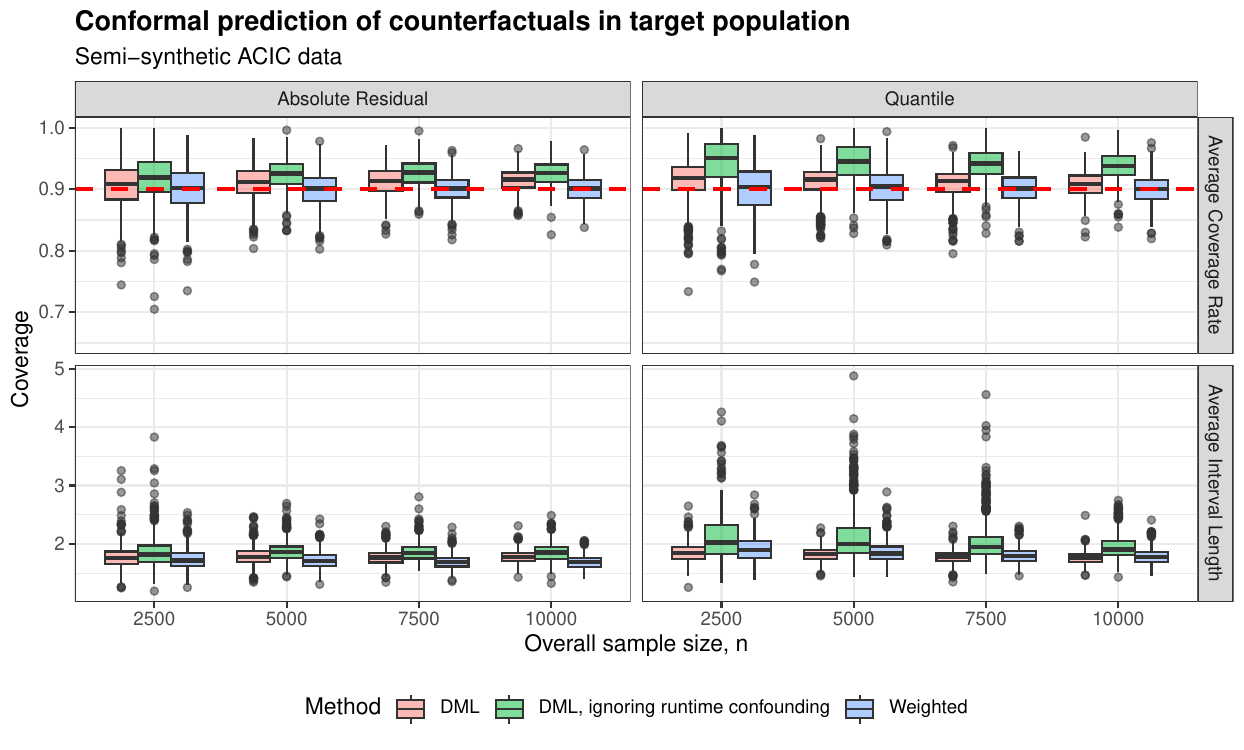}
    \vspace{-1.2em}
    \caption{Performance of proposed methods on semi-synthetic ACIC data.}
    \label{fig:sim-results}
    \vspace{-1em}
\end{figure}

\textbf{Results:} Figure \ref{fig:sim-results} displays the results of our exercise for the moderate runtime confounding scenario. Results for the mild and severe scenarios are qualitatively similar and reported in Appendix \ref{ap:extra-data-app-results}. Similar to our numerical experiments in Section \ref{sec:num-exp}, we see that our proposed DML procedure and the weighted conformal approach based on our derived weights $w_a(\bm O)$ both attain approximately valid coverage. Naively applying \cite{yang2024doubly} and ignoring runtime confounding continues to lead to miscoverage that is most pronounced when using quantile scores. Along with possessing the largest coverage bias, the naive approach consistently produces the widest prediction intervals, further demonstrating the consequences that can arise from ignoring runtime confounding.  Notably, the weighted procedure based on our derived weights $\hat w_a(\bm O)$ performs well over all sample sizes considered. We suspect the slight coverage bias for our proposed DML method arises from relative complexity in the underlying conditional score functions $q_a$ and $m_a$, recalling the weighted procedure does not require estimates of these functions. Consistent with Theorem \ref{thm:dr-thm}, the proposed DML procedure  attains approximately valid coverage throughout, since accurate estimation of the nuisance functions comprising $\hat w_a(\bm O)$ partially protects against inaccurate estimation of $q_a$ and $m_a$. 
\vspace{-2.5em}
\section{Discussion}
\label{sec:discussion}

We developed computationally and statistically efficient methods to construct prediction intervals for counterfactual outcomes under runtime confounding, a setting that involves both treatment-outcome confounding and covariate shift between source and target populations. Our approach uses a multiply robust debiased machine learning estimator of the required conformity score quantile, enabling the resulting prediction intervals to achieve desired coverage rates under modest nuisance learning requirements. Our theoretical results show this coverage is achieved more rapidly as a function of $n$ than with standard plug-in methods, and numerical experiments identify numerous scenarios where our method displays superior or comparable performance to standard approaches. Additionally, we provided valid loss functions for performing counterfactual quantile regression in runtime confounding settings, and a weighted conformal prediction method that effectively addressed runtime confounding bias throughout our numerical experiments. Both the proposed DML method and the weighted procedure consistently outperform a state-of-the-art DML procedure that ignores runtime confounding in our numerical experiments, highlighting the need to address runtime confounding in practice.

A limitation of our approach is that the validity of our method relies on the causal and transportability  Assumptions \ref{as:assumption-positivty}-\ref{as:source-pos}, which are untestable. The procedure may fail if, for example, there is an unmeasured confounder of the treatment-outcome relationship or if the source and target populations differ in ways not captured by the observed covariates $\bm V$. 
Future work could extend our framework to several important areas. 
One direction is developing formal sensitivity analyses to quantify how prediction intervals and coverage rates are affected by violations of the core independence Assumptions \ref{as:assumption-unconfoundedness}-\ref{as:source-exch}. Extending the sensitivity analysis framework from \cite{zeng2025efficient}, who focused on ATE estimation under runtime confounding, could serve as a promising avenue forward. Additionally, our framework could be extended to support continuous treatments  in runtime confounding settings \citep{schroder2025conformal} and survival outcomes \citep{candes2023conformalized}.

\clearpage

\acks{This work was funded by National Science Foundation grant NSF DMS 1810829.}

\bibliography{sources}

@article{alaa2023conformal,
  title={Conformal meta-learners for predictive inference of individual treatment effects},
  author={Alaa, Ahmed M and Ahmad, Zaid and van der Laan, Mark},
  journal={Advances in neural information processing systems},
  volume={36},
  pages={47682--47703},
  year={2023}
}

@article{angelopoulos2023conformal,
  title={Conformal prediction: A gentle introduction},
  author={Angelopoulos, Anastasios N and Bates, Stephen},
  journal={Foundations and Trends in Machine Learning},
  volume={16},
  number={4},
  pages={494--591},
  year={2023},
  publisher={Emerald Publishing Limited}
}

@article{angelopoulos2024theoretical,
  title={Theoretical foundations of conformal prediction},
  author={Angelopoulos, Anastasios N and Barber, Rina Foygel and Bates, Stephen},
  journal={arXiv preprint arXiv:2411.11824},
  year={2024}
}

@article{barber2023conformal,
  title={Conformal prediction beyond exchangeability},
  author={Barber, Rina Foygel and Candes, Emmanuel J and Ramdas, Aaditya and Tibshirani, Ryan J},
  journal={The Annals of Statistics},
  volume={51},
  number={2},
  pages={816--845},
  year={2023},
  publisher={Institute of Mathematical Statistics}
}

@article{bareinboim2016causal,
  title={Causal inference and the data-fusion problem},
  author={Bareinboim, Elias and Pearl, Judea},
  journal={Proceedings of the National Academy of Sciences},
  volume={113},
  number={27},
  pages={7345--7352},
  year={2016},
  publisher={National Academy of Sciences}
}

@article{bica2022transfer,
  title={Transfer learning on heterogeneous feature spaces for treatment effects estimation},
  author={Bica, Ioana and van der Schaar, Mihaela},
  journal={Advances in Neural Information Processing Systems},
  volume={35},
  pages={37184--37198},
  year={2022}
}

@article{bleier2020consumer,
  title={Consumer privacy and the future of data-based innovation and marketing},
  author={Bleier, Alexander and Goldfarb, Avi and Tucker, Catherine},
  journal={International Journal of Research in Marketing},
  volume={37},
  number={3},
  pages={466--480},
  year={2020},
  publisher={Elsevier}
}

@article{boyer2025estimating,
  title={Estimating and evaluating counterfactual prediction models},
  author={Boyer, Christopher B and Dahabreh, Issa J and Steingrimsson, Jon A},
  journal={Statistics in Medicine},
  volume={44},
  number={23-24},
  pages={e70287},
  year={2025},
  publisher={Wiley Online Library}
}

@article{candes2023conformalized,
  title={Conformalized survival analysis},
  author={Candes, Emmanuel and Lei, Lihua and Ren, Zhimei},
  journal={Journal of the Royal Statistical Society Series B: Statistical Methodology},
  volume={85},
  number={1},
  pages={24--45},
  year={2023},
  publisher={Oxford University Press US}
}

@inproceedings{chouldechova2018case,
  title={A case study of algorithm-assisted decision making in child maltreatment hotline screening decisions},
  author={Chouldechova, Alexandra and Benavides-Prado, Diana and Fialko, Oleksandr and Vaithianathan, Rhema},
  booktitle={Conference on fairness, accountability and transparency},
  pages={134--148},
  year={2018},
  organization={PMLR}
}

@article{collins2023prediction,
  title={Prediction models should contain predictors known at the moment of intended use},
  author={Collins, Gary S and Dhiman, Paula},
  journal={Aging Clinical and Experimental Research},
  volume={35},
  number={12},
  pages={3243--3244},
  year={2023},
  publisher={Springer}
}

@article{coston2020counterfactual,
  title={Counterfactual predictions under runtime confounding},
  author={Coston, Amanda and Kennedy, Edward and Chouldechova, Alexandra},
  journal={Advances in neural information processing systems},
  volume={33},
  pages={4150--4162},
  year={2020}
}

@article{chernozhukov2018double,
  title={Double/debiased machine learning for treatment and structural parameters},
  author={Chernozhukov, Victor and Chetverikov, Denis and Demirer, Mert and Duflo, Esther and Hansen, Christian and Newey, Whitney and Robins, James},
  journal={The Econometrics Journal},
  pages={C1--C68},
  year={2018},
  publisher={JSTOR}
}

@article{degtiar2023review,
  title={A review of generalizability and transportability},
  author={Degtiar, Irina and Rose, Sherri},
  journal={Annual Review of Statistics and Its Application},
  volume={10},
  number={1},
  pages={501--524},
  year={2023},
  publisher={Annual Reviews}
}

@article{deschepper2025literature,
  title={A literature-based approach to predict continuous hospital length of stay in adult acute care patients using admission variables: A single university center experience},
  author={Deschepper, Mieke and De Smedt, Chlo{\"e} and Colpaert, Kirsten},
  journal={International Journal of Medical Informatics},
  volume={193},
  pages={105678},
  year={2025},
  publisher={Elsevier}
}

@article{carvalho2019assessing,
  title={Assessing treatment effect variation in observational studies: Results from a data challenge},
  author={Carvalho, Carlos and Feller, Avi and Murray, Jared and Woody, Spencer and Yeager, David},
  journal={Observational Studies},
  volume={5},
  number={2},
  pages={21--35},
  year={2019},
  publisher={University of Pennsylvania Press}
}

@article{colnet2024causal,
  title={Causal inference methods for combining randomized trials and observational studies: a review},
  author={Colnet, B{\'e}n{\'e}dicte and Mayer, Imke and Chen, Guanhua and Dieng, Awa and Li, Ruohong and Varoquaux, Ga{\"e}l and Vert, Jean-Philippe and Josse, Julie and Yang, Shu},
  journal={Statistical science},
  volume={39},
  number={1},
  pages={165--191},
  year={2024},
  publisher={Institute of Mathematical Statistics}
}

@article{fischer2024bridging,
  title={Bridging the gap: Towards an expanded toolkit for AI-driven decision-making in the public sector},
  author={Fischer-Abaigar, Unai and Kern, Christoph and Barda, Noam and Kreuter, Frauke},
  journal={Government Information Quarterly},
  volume={41},
  number={4},
  pages={101976},
  year={2024},
  publisher={Elsevier}
}

@article{fontana2023conformal,
  title={Conformal prediction: a unified review of theory and new challenges},
  author={Fontana, Matteo and Zeni, Gianluca and Vantini, Simone},
  journal={Bernoulli},
  volume={29},
  number={1},
  pages={1--23},
  year={2023},
  publisher={Bernoulli Society for Mathematical Statistics and Probability}
}

@inproceedings{gao2025bridging,
  title={Bridging Fairness and Efficiency in Conformal Inference: A Surrogate-Assisted Group-Clustered Approach},
  author={Gao, Chenyin and Gilbert, Peter B and Han, Larry},
  booktitle={Forty-second International Conference on Machine Learning},
year={2025}
}

@article{graham2024towards,
  title={Towards a unified theory for semiparametric data fusion with individual-level data},
  author={Graham, Ellen and Carone, Marco and Rotnitzky, Andrea},
  journal={arXiv preprint arXiv:2409.09973},
  year={2024}
}

@article{han2023multiply,
  title={Multiply robust federated estimation of targeted average treatment effects},
  author={Han, Larry and Shen, Zhu and Zubizarreta, Jose},
  journal={Advances in Neural Information Processing Systems},
  volume={36},
  pages={70453--70482},
  year={2023}
}

@article{han2025federated,
  title={Federated adaptive causal estimation (face) of target treatment effects},
  author={Han, Larry and Hou, Jue and Cho, Kelly and Duan, Rui and Cai, Tianxi},
  journal={Journal of the American Statistical Association},
  pages={1--14},
  year={2025},
  publisher={Taylor \& Francis}
}

@article{kallus2025role,
  title={On the role of surrogates in the efficient estimation of treatment effects with limited outcome data},
  author={Kallus, Nathan and Mao, Xiaojie},
  journal={Journal of the Royal Statistical Society Series B: Statistical Methodology},
  volume={87},
  number={2},
  pages={480--509},
  year={2025},
  publisher={Oxford University Press UK}
}

@article{kallus2024localized,
  title={Localized debiased machine learning: Efficient inference on quantile treatment effects and beyond},
  author={Kallus, Nathan and Mao, Xiaojie and Uehara, Masatoshi},
  journal={Journal of Machine Learning Research},
  volume={25},
  number={16},
  pages={1--59},
  year={2024}
}

@article{kennedy2020efficient,
  title={Efficient nonparametric causal inference with missing exposure information},
  author={Kennedy, Edward H},
  journal={The international journal of biostatistics},
  volume={16},
  number={1},
  year={2020},
  publisher={De Gruyter}
}

@article{kennedy2024semiparametric,
  title={Semiparametric doubly robust targeted double machine learning: a review},
  author={Kennedy, Edward H},
  journal={Handbook of Statistical Methods for Precision Medicine},
  pages={207--236},
  year={2024},
  publisher={Chapman and Hall/CRC}
}

@article{keogh2024prediction,
  title={Prediction under interventions: evaluation of counterfactual performance using longitudinal observational data},
  author={Keogh, Ruth H and Van Geloven, Nan},
  journal={Epidemiology},
  volume={35},
  number={3},
  pages={329--339},
  year={2024},
  publisher={LWW}
}

@article{koenker1978regression,
  title={Regression quantiles},
  author={Koenker, Roger and Bassett Jr, Gilbert},
  journal={Econometrica: journal of the Econometric Society},
  pages={33--50},
  year={1978},
  publisher={JSTOR}
}

@article{lei2021conformal,
  title={Conformal inference of counterfactuals and individual treatment effects},
  author={Lei, Lihua and Cand{\`e}s, Emmanuel J},
  journal={Journal of the Royal Statistical Society Series B: Statistical Methodology},
  volume={83},
  number={5},
  pages={911--938},
  year={2021},
  publisher={Oxford University Press}
}

@article{little2000causal,
  title={Causal effects in clinical and epidemiological studies via potential outcomes: concepts and analytical approaches},
  author={Little, Roderick J and Rubin, Donald B},
  journal={Annual review of public health},
  volume={21},
  number={1},
  pages={121--145},
  year={2000},
  publisher={Annual Reviews 4139 El Camino Way, PO Box 10139, Palo Alto, CA 94303-0139, USA}
}

@inproceedings{liu2024multi,
  title={Multi-Source Conformal Inference Under Distribution Shift},
  author={Liu, Yi and Levis, Alexander and Normand, Sharon-Lise and Han, Larry},
  booktitle={International Conference on Machine Learning},
  pages={31344--31382},
  year={2024},
  organization={PMLR}
}

@incollection{musen2021clinical,
  title={Clinical decision-support systems},
  author={Musen, Mark A and Middleton, Blackford and Greenes, Robert A},
  booktitle={Biomedical informatics: computer applications in health care and biomedicine},
  pages={795--840},
  year={2021},
  publisher={Springer}
}

@inproceedings{park2025semiparametric,
  title={Semiparametric conformal prediction},
  author={Park, Ji Won and Cho, Kyunghyun},
  booktitle={International Conference on Artificial Intelligence and Statistics},
  pages={3880--3888},
  year={2025},
  organization={PMLR}
}

@article{romano2019conformalized,
  title={Conformalized quantile regression},
  author={Romano, Yaniv and Patterson, Evan and Candes, Emmanuel},
  journal={Advances in neural information processing systems},
  volume={32},
  year={2019}
}

@article{rojas2018invariant,
  title={Invariant models for causal transfer learning},
  author={Rojas-Carulla, Mateo and Sch{\"o}lkopf, Bernhard and Turner, Richard and Peters, Jonas},
  journal={Journal of Machine Learning Research},
  volume={19},
  number={36},
  pages={1--34},
  year={2018}
}

@article{rubin2005causal,
  title={Causal inference using potential outcomes: Design, modeling, decisions},
  author={Rubin, Donald B},
  journal={Journal of the American statistical Association},
  volume={100},
  number={469},
  pages={322--331},
  year={2005},
  publisher={Taylor \& Francis}
}

@article{rosenbaum1983central,
  title={The central role of the propensity score in observational studies for causal effects},
  author={Rosenbaum, Paul R and Rubin, Donald B},
  journal={Biometrika},
  volume={70},
  number={1},
  pages={41--55},
  year={1983},
  publisher={Oxford University Press}
}

@inproceedings{schroder2025conformal,
  title={Conformal Prediction for Causal Effects of Continuous Treatments},
  author={Schr{\"o}der, Maresa and Frauen, Dennis and Schweisthal, Jonas and Hess, Konstantin and Melnychuk, Valentyn and Feuerriegel, Stefan},
  booktitle={The Thirty-ninth Annual Conference on Neural Information Processing Systems},
  year={2025}
}

@article{shyr2025multi,
  title={Multi-study R-learner for estimating heterogeneous treatment effects across studies using statistical machine learning},
  author={Shyr, Cathy and Ren, Boyu and Patil, Prasad and Parmigiani, Giovanni},
  journal={Biostatistics},
  volume={26},
  number={1},
  pages={kxaf040},
  year={2025},
  publisher={Oxford University Press}
}

@article{tibshirani2019conformal,
  title={Conformal prediction under covariate shift},
  author={Tibshirani, Ryan J and Foygel Barber, Rina and Candes, Emmanuel and Ramdas, Aaditya},
  journal={Advances in neural information processing systems},
  volume={32},
  year={2019}
}

@book{tsiatis2006semiparametric,
  title={Semiparametric theory and missing data},
  author={Tsiatis, Anastasios A},
  volume={4},
  year={2006},
  publisher={Springer}
}

@book{van2000asymptotic,
  title={Asymptotic Statistics},
  author={van der Vaart, AW},
  volume={3},
  year={2000},
  publisher={Cambridge University Press}
}

@article{voter2025counterfactual,
  title={Counterfactual prediction from machine learning models: transportability and joint analysis for model development and evaluation using multi-source data},
  author={Voter, Sarah C and Dahabreh, Issa J and Boyer, Christopher B and Rahbar, Habib and Kontos, Despina and Steingrimsson, Jon A},
  journal={Diagnostic and Prognostic Research},
  volume={9},
  number={1},
  pages={22},
  year={2025},
  publisher={Springer}
}

@book{vovk2005algorithmic,
  title={Algorithmic learning in a random world},
  author={Vovk, Vladimir and Gammerman, Alexander and Shafer, Glenn},
  volume={29},
  year={2005},
  publisher={Springer}
}

@article{yang2024doubly,
  title={Doubly robust calibration of prediction sets under covariate shift},
  author={Yang, Yachong and Kuchibhotla, Arun Kumar and Tchetgen Tchetgen, Eric},
  journal={Journal of the Royal Statistical Society Series B: Statistical Methodology},
  volume={86},
  number={4},
  pages={943--965},
  year={2024},
  publisher={Oxford University Press US}
}

@article{yeager2019national,
  title={A national experiment reveals where a growth mindset improves achievement},
  author={Yeager, David S and Hanselman, Paul and Walton, Gregory M and Murray, Jared S and Crosnoe, Robert and Muller, Chandra and Tipton, Elizabeth and Schneider, Barbara and Hulleman, Chris S and Hinojosa, Cintia P and others},
  journal={Nature},
  volume={573},
  number={7774},
  pages={364--369},
  year={2019},
  publisher={Nature Publishing Group}
}

@article{zeng2025efficient,
  author  = {Zeng, Zhenghao and Kennedy, Edward H. and Bodnar, Lisa M. and Naimi, Ashley I.},
  title   = {Efficient Generalization and Transportation},
  journal = {Statistical Science},
  volume  = {40},
  number  = {3},
  pages   = {495--514},
  year    = {2025},
  month   = aug,
}

\clearpage

\appendix

\section{Proofs}
\label{sec:proofs}

\subsection{Lemmas}

\begin{lemma}
\label{lem:wgt-lemma}
    Let $h(Y(a),\bm V)$ be a generic square integrable function. Under Assumptions \ref{as:assumption-positivty}-\ref{as:source-pos} we have that
    \[
    \E[h(Y(a),\bm V) | S=0] = \E\left(\mathbb{I}(A=a) S h(Y,\bm V) \frac{1-\kappa(\bm V)}{g_a(\bX) \kappa(\bm V)}\right).
    \]
\end{lemma}
To prove Lemma \ref{lem:wgt-lemma}---which we in turn leverage in the proofs of Proposition \ref{prop:wgtd-quant} and Theorem \ref{thm:identification}---notice
\begin{align*}
    \E\left(\mathbb{I}(A=a) S h(Y(a),\bm V) \frac{1-\kappa(\bm V)}{g_a(\bX) \kappa(\bm V)}\right) 
    &=
    \E\left( \E[\mathbb{I}(A=a) S h(Y(a),\bm V) | \bX] \frac{1-\kappa(\bm V)}{g_a(\bX) \kappa(\bm V)}\right) \\
    &=
   \E\left( \E[ h(Y(a),\bm V) | \bX] g_a(\bX) \PP(S=1|\bX) \frac{1-\kappa(\bm V)}{g_a(\bX) \kappa(\bm V)}\right) \\
   &=
   \E\left( \E[ h(Y(a),\bm V) | \bX]  \PP(S=1|\bX) \frac{1-\kappa(\bm V)}{\kappa(\bm V)}\right) \\
   &=
   \E\left( \E\left\{ S h(Y(a),\bm V) \cdot   \frac{1-\kappa(\bm V)}{\kappa(\bm V)} \bigg | \bX \right\}\right) \\
   &=
   \E\left(S h(Y(a),\bm V) \cdot   \frac{1-\kappa(\bm V)}{\kappa(\bm V)} \right) \\
   &=
   \E\left( \E\left\{ S h(Y(a),\bm V) \cdot   \frac{1-\kappa(\bm V)}{\kappa(\bm V)} \bigg | \bm V \right\}\right) \\
   &=
     \E\left( h(Y(a),\bm V) |  S=0 \right). 
\end{align*}

\subsection{Proof of Proposition \ref{prop:wgtd-quant}}

Let $Q_{a,\alpha}(\bm v)$ denote the $1-\alpha$ quantile of $Y(a)$ conditional on $\bm V=\bm v$, and note $Q_{a,\alpha}(\bm v)$ satisfies \citep{koenker1978regression}
\[
\PP(Y(a) \leq Q_{a,\alpha}(\bm V) | \bm V, S=0) = 1-\alpha.
\]
Letting $\rho_{\alpha}(x) := \alpha |x|\mathbb{I}(x>0) + (1-\alpha)|x|\mathbb{I}(x\leq 0)$ denote the pinball loss function, notice $Q_{a,\alpha}(\bm V)$ additionally satisfies
\[
Q_{a,\alpha}(\bm V) = \argmin_{\tilde Q_{a,\alpha}}  \E[\rho_{\alpha}\{Y(a) - Q_{a,\alpha}(\bm V)\}|S=0]
\]
For brevity, let $h_{\alpha}(\bm V,Y(a);Q) := \rho_{\alpha}\{Y(a) - Q(\bm V)\}$, and notice 
\[
\E[h_{\alpha}(\bm V,Y(a);Q) | S=0] = \E\left[\mathbb{I}(A=a) S h_\alpha(\bm V, Y(a) ; Q) \frac{1-\kappa(\bm V)}{g_a(\bm X) \kappa(\bm V)} \right]
\]
by Lemma \ref{lem:wgt-lemma}, which implies
\[
Q_{a,\alpha}(\bm V) = \argmin_{\tilde Q_{a,\alpha}}   \E\left[\mathbb{I}(A=a) S h_\alpha(\bm V, Y(a) ; \tilde Q_{a,\alpha}) \frac{1-\kappa(\bm V)}{g_a(\bm X) \kappa(\bm V)} \right],
\]
proving Proposition \ref{prop:wgtd-quant} and suggesting one can estimate conditional quantiles of $Y(a) | \bm V$ through a simple reweighting of the conventional quantile regression loss function.

\subsection{Proof of Theorem \ref{thm:identification}}

We begin by proving (\ref{eq:thm1-gcomp}). For compactness, let $R_a := R_a(Y,\bm V)$ and notice
\begin{align*}
    \PP(R_{a} \leq r_{a,\alpha} | S=0) &=
    \E(\mathbb{I}(R_{a} \leq r_{a,\alpha}) | S=0) \\
    &=
    \E[ \E(\mathbb{I}(R_{a} \leq r_{a,\alpha}) | \bX, S=0) | S=0 ]
    \\
    &=
    \E[ \E(\mathbb{I}(R_{a} \leq r_{a,\alpha}) | \bX, S=1) | S=0 ] \\
    &=
    \E \{ \E[ \E(\mathbb{I}(R_{a} \leq r_{a,\alpha}) | \bX, A=a, S=1) | \bm V,S=1] | S=0 \}
\end{align*}
Above, line 3 holds by Assumption \ref{as:source-exch}, recalling $\bm V \subset \bX$, and line 4 holds due to Assumptions \ref{as:assumption-unconfoundedness} and \ref{as:source-exch}. The desired result holds by recalling the definitions of $q_a(r_{a,\alpha},\bm X)$ and $m_a(r_{a,\alpha},\bm V)$.
\\[1.2em]
Letting $h(Y(a),\bm V):=I(R_a(Y,\bm V)\leq r_{a,\alpha})$, Equation \eqref{eq:thm1-ipw} immediately follows by Lemma \ref{lem:wgt-lemma}.

\subsection{Proof of Theorem \ref{thm:eif-thm}}

\newcommand{\pwderiv}{\frac{d}{d\varepsilon}}
\newcommand{\evalzero}{\bigg |_{\varepsilon=0}}
\newcommand{\EPPe}{\mathbb{E}_{\mathbb{P}_{\varepsilon}}}
\newcommand{\PPe}{\mathbb{P}_\varepsilon}
\newcommand{\EP}{\mathbb{E}_\mathbb{P}}

Suppose $\bm O \sim \PP$, and let $\{\PP_\varepsilon: \varepsilon \in [0,1)\}$ be a generic regular parametric submodel containing the true data-generating distribution at $\varepsilon=0: \PP_0=\PP$.
Recall that an \textit{influence curve} for a pathwise differentiable functional $\Psi(\PP)$ is a function $\chi(\bm O,\PP)$ which satisfies
\begin{equation}
\label{eq:eif-definition}
   \pwderiv \Psi(\PP_\varepsilon) \evalzero = \E_\PP [\chi(\bm O,\PP) u(\bm O)], 
\end{equation}
for any parametric submodel, where $\E_\PP[\chi(\bm O,\PP)]=0, \ \text{Var}_\PP(\chi(\bm O,\PP)) < \infty$ and $u(\bm O) = \pwderiv \log \PP_\varepsilon |_{\varepsilon=0}$ is the score function for the parametric submodel evaluated at $\varepsilon=0$ \citep{tsiatis2006semiparametric}. For generic $Q,W$, let $u_{Q|W}$ be the conditional score of $Q|W$, $u_{Q,W}$ be the score function for the joint distribution of $Q$ and $W$, $u_W$ be the score function for $W$, and note that $u_{Q,W}=u_{Q|W} + u_W$. The tangent space is defined as the closed linear span of scores for all possible parametric submodels, and the \textit{efficient influence curve} is the unique influence function belonging to the tangent space.

Analogous to \cite{liu2024multi}---whose approach we follow---our strategy to find the efficient influence curve for $r_{a,\alpha}(\PP)$ is to begin by differentiating the identifying expression (\ref{eq:thm1-gcomp}) induced by a generic parametric submodel $\PP_\varepsilon$, where we will ultimately rearrange the resulting terms to arrive at an expression for $\pwderiv r_{a,\alpha}(\PP_\varepsilon)|_{\varepsilon=0}$. Notice we have
\begin{align*}
    0 &= \pwderiv  \EPPe \{ \ \EPPe[ \ \EPPe(\mathbb{I}(R_{a}(Y,\bm V) \leq r_{a,\alpha}(\PPe)) \ | 
   \ \bX, A=a, S=1) \ | \ \bm V, S=1 \ ] \ | \  S=0 \ \} \evalzero \\
   &= \pwderiv  \EPPe \{ \ \EP[ \ \EP(\mathbb{I}(R_{a}(Y,\bm V) \leq r_{a,\alpha}(\PP)) \ | 
   \ \bX, A=a, S=1) \ | \ \bm V, S=1 \ ] \ | \  S=0 \ \} \evalzero 
   \\ 
   &+
   \pwderiv  \EP \{ \ \EPPe[ \ \EP(\mathbb{I}(R_{a}(Y,\bm V) \leq r_{a,\alpha}(\PP)) \ | 
   \ \bX, A=a, S=1) \ | \ \bm V, S=1 \ ] \ | \  S=0 \ \} \evalzero \\ 
   &+
   \pwderiv  \EP \{ \ \EP[ \ \EPPe(\mathbb{I}(R_{a}(Y,\bm V) \leq r_{a,\alpha}(\PP)) \ | 
   \ \bX, A=a, S=1) \ | \ \bm V, S=1 \ ] \ | \  S=0 \ \} \evalzero \\
   &+
   \pwderiv  \EP \{ \ \EP[ \ \EP(\mathbb{I}(R_{a}(Y,\bm V) \leq r_{a,\alpha}(\PPe)) \ | 
   \ \bX, A=a, S=1) \ | \ \bm V, S=1 \ ] \ | \  S=0 \ \} \evalzero \\
   &= \text{I} + \text{II} + \text{III} + \text{IV}
\end{align*}
Focusing on the first term and recalling the definition of $m_a(r,\bm V)$ we have 
\begin{align*}
\text{I} 
&= 
\pwderiv \EPPe \{ m_a(r_{a,\alpha}(\PP),\bm V) | S=0 \} \evalzero
\\
&=
\EP \{(m_a(r_{a,\alpha}(\PP),\bm V) - (1-\alpha)) u_{\bm V | S=0} | S=0  \} \\
&=
\EP \left\{\frac{1-S}{\PP(S=0)}(m_a(r_{a,\alpha}(\PP),\bm V) - (1-\alpha)) u_{\bm V | S}  \right\} \\
&=
\EP \left\{\frac{1-S}{\PP(S=0)}(m_a(r_{a,\alpha}(\PP),\bm V) - (1-\alpha)) u(\bm O)  \right\}
\end{align*}
Above, we are able to add in $u_{S}$ since $(1-S)(m_a(r_{a,\alpha}(\PP),\bm V) - (1-\alpha))$ has mean zero given $S$, and in turn can then add in $u_{Y,A,\bm U|\bm V,S}$, which is mean zero given $\bm V$ and $S$. 
\\ \\
For the second term, recalling $\bX=(\bm V, \bm U)$ and that $q_a(r_{a,\alpha}(\PP),\bX) = \E_{\PP}[\II(R_a(Y,\bm V) \leq r_{a,\alpha}(\PP) | \bX, A=a, S=1]$, notice
\begin{align*}
   \text{II} &= \pwderiv  \EP \{ \EPPe[ q_a(r_{a,\alpha}(\PP),\bX) | \bm V] | S=0   \} \\
   &=
  \EP \{ \EP[ q_a(r_{a,\alpha}(\PP),\bX) - m_a(r_{a,\alpha}(\PP),\bm V))u_{\bm U|\bm V,S=1} | \bm V, S=1] | S=0   \} \\
  &=
\EP \left\{ \frac{1-S}{\PP(S=0)} \EP[ q_a(r_{a,\alpha}(\PP),\bX) - m_a(r_{a,\alpha}(\PP),\bm V))u_{\bm U|\bm V,S=1}  | \bm V, S=1] \right\}  \\
&=
\EP \left\{ \frac{1-S}{\PP(S=0)} \EP\left[ 
\frac{S}{\kappa(\bm V)}
\{q_a(r_{a,\alpha}(\PP),\bX) - m_a(r_{a,\alpha}(\PP),\bm V)\}u_{\bm U|\bm V,S} \right]   \right\} \\
&=
\EP \left\{ \frac{1-\kappa(\bm V)}{\PP(S=0)} 
\frac{S}{\kappa(\bm V)}
\{q_a(r_{a,\alpha}(\PP),\bX) - m_a(r_{a,\alpha}(\PP),\bm V)\}u_{\bm U|\bm V,S}    \right\} \\
&=
\EP \left\{ \frac{1-\kappa(\bm V)}{\PP(S=0)} 
\frac{S}{\kappa(\bm V)}
\{q_a(r_{a,\alpha}(\PP),\bX) - m_a(r_{a,\alpha}(\PP),\bm V)\}u(\bm O)    \right\} 
\end{align*} \\
Similar to term I, on the final line we are able to add in $u_{\bm V,S}$ since this term is mean zero given $\bm V$ and $S$. Recalling $\bX=(\bm V,\bm U)$, we can then add in $u_{Y,A|\bX, S}$ following the same logic used in term I.
\\ \\
For the third term, following similar logic we have
{\footnotesize
\begin{align*}
   \text{III} &= \EP \{ \EP [ \EP[  (\II(R_a(Y,\bm V) \leq r_{a,\alpha}(\PP)) - q_a(r_{a,\alpha}(\PP),\bX))u_{Y|\bX,A=a,S=1}|\bX,A=a] | \bm V, S=1\ | S=0 \} \\
   &=
 \EP \{ \EP [ \EP\left[\frac{\mathbb{I}(A=a)S}{\PP(A=a,\bX,S=1)}  (\II(R_a(Y,\bm V) \leq r_{a,\alpha}(\PP)) - q_a(r_{a,\alpha}(\PP),\bX))u_{Y|\bX,A,S}\right] | \bm V, S=1\ | S=0 \}  \\
 &=
 \EP\left[ \frac{1-\kappa(\bm V)}{\PP(S=0)}\frac{\mathbb{I}(A=a)S}{\PP(A=a,\bX,S=1)}\{\II(R_a(Y,\bm V) - q_a(r_{a,\alpha}(\PP),\bX) \}u(\bm O) \right]
\end{align*}}
\\ 
Finally, for the fourth term we have that
\begin{align*}
   \text{IV} &= \pwderiv \EP[m_a(r_{a,\alpha} 
 (\PPe)) | S=0 ]\evalzero \\
 &\propto \pwderiv r_{a,\alpha}(\PP_\varepsilon) \evalzero
\end{align*}
Above, we can safely ignore the proportionality constant, since influence curves are by construction mean zero and we intend to use the resulting influence curve to form estimating equation estimators of $r_{a,\alpha}(\PP)$, whose solutions are invariant to scaling. Note that if we wished to perform a one-step bias correction, we would need to incorporate this proportionality constant. Since we do not wish to perform statistical inference on $r_{a,\alpha}$, and instead simply require an efficient estimate of this quantity, there are no costs incurred by avoiding estimation of this constant.

Re-combining I, II, III and IV, and solving for $\pwderiv r_{a,\alpha}(\PP)|_{\varepsilon=0}$, we have that 
\begin{align*}
\pwderiv r_{a,\alpha}(\PP)\evalzero &\propto 
    \EP \left\{\frac{1-S}{\PP(S=0)}(m_a(r_{a,\alpha}(\PP),\bm V) - (1-\alpha)) u(\bm O)  \right\} \\
    &+
    \EP \left\{ \frac{1-\kappa(\bm V)}{\PP(S=0)} 
\frac{S}{\kappa(\bm V)}
\{q_a(r_{a,\alpha}(\PP),\bX) - m_a(r_{a,\alpha}(\PP),\bm V)\}u(\bm O)    \right\} \\
&+
\EP\left[ \frac{1-\kappa(\bm V)}{\PP(S=0)}\frac{\mathbb{I}(A=a)S}{\PP(A=a,\bX,S=1)}\{\II(R_a(Y,\bm V) - q_a(r_{a,\alpha}(\PP),\bX) \}u(\bm O) \right]
\end{align*}
Noting each term above is mean zero, 
\begin{align*}
    \pwderiv r_{a,\alpha}(\PP)\evalzero  &\propto \EP \bigg[ 
    \bigg\{ (1-S)(m_a(r_{a,\alpha}(\PP),\bm V) - (1-\alpha))    +
\frac{S(1-\kappa(\bm V))}{\kappa(\bm V)}
\{q_a(r_{a,\alpha}(\PP),\bX) - m_a(r_{a,\alpha}(\PP),\bm V) \}\\
&+ 
\frac{\mathbb{I}(A=a)S)1-\kappa(\bm V))}{\PP(A=a,\bX,S=1)}\{\II(R_a(Y,\bm V) - q_a(r_{a,\alpha}(\PP),\bX) \}
\bigg\}u(\bm O)\bigg] \\
&= \EP[\chi_a(\bm O,\PP; m_a,r_a,g_a,\kappa)u(\bm O)]
\end{align*}
where we additionally omit the proportionality constant $\PP(S=0)^{-1}$ initially appearing in each of the three terms above for brevity. It is then straightforward to verify that $\chi_a(\bm O)$ is an element of the tangent space.

Recalling the definition of an EIC in (\ref{eq:eif-definition}), since $\chi_a(\bm O,\PP; m_a,r_a,g_a,\kappa)$ is mean zero, we conclude that the efficient influence curve for $r_{a,\alpha}(\PP)$ is proportional to $\chi_a(\bm O,\PP; m_a,r_{a,\alpha},g_a,\kappa)$. 

\subsection{Proof of Theorem \ref{thm:dr-thm}}
\newcommand{\hatra}{\hat r_{a,\alpha}}

Suppose that $\hat \eta_a = (\hat q_a, \hat m_a, \hat \kappa, \hat g)$ is obtained from a separate sample independent from $\bm O_i$, and assume there exists some small $\varepsilon>0$ such that $\hat \kappa(\bm V) \in (\varepsilon, 1-\varepsilon)$, $\hat g_a(\bX) \in (\varepsilon, 1-\varepsilon)$, and $\PP(S=1|\bX) \in (\varepsilon,1-\varepsilon)$ almost surely. 

We aim to show that 
\begin{equation}
\label{eq:cov-rate}
    \PP(Y(a) \in \hat C_a(\bm V) | S=0) = 1-\alpha + O_\PP(1/\sqrt n \ + \ R_n),    
\end{equation}
where $R_n = \sup_{r}|| \hat q_a(r,\cdot) - q_a(r,\cdot)|| \cdot ||\hat g_a - g_a || + \sup_r|| \hat m_a(r,\cdot) - m_a(r,\cdot) || \cdot ||\hat \kappa - \kappa ||$. Such a construction allows for one to quantify conditions on nuisance function estimation rates such that the above coverage slack is of order $O_\PP(1/\sqrt n)$.

To achieve this, we will  
\begin{enumerate}
    \item Show $\PP(Y(a) \in \hat C_a(\bm V) | S=0) -  (1-\alpha) = \E[\chi_a(\bm O,\hatra;\eta)]/\PP(S=0)$
    \item Decompose $\E[\chi_a(\bm O,\hatra\;\eta)]$ into a term whose asymptotic behavior is dominated by $\E(\chi_a(\hatra,\hat \eta) - \chi_a(\hatra, \eta))$
    \item Show that for any $r$, the difference $\E(\chi_a(r,\hat \eta) - \chi_a(r, \eta))$ satisfies the product bias structure specified in Theorem \ref{thm:dr-thm}
    \item Take the supremum of this bias structure over all $r$ to bound $\E(\chi_a(\hatra,\hat \eta) - \chi_a(\hatra, \eta))$
\end{enumerate}
To begin,  notice 
\begin{align}
    \PP(Y(a) \in \hat C_a(\bm V) | S=0) - (1-\alpha) &= \PP(R_a(Y(a),\bm V) \leq \hat r_{a,\alpha})  | S=0 ) - (1-\alpha) \nonumber \\
    &= 
    \E[\chi_a(\bm O,\hatra;\eta)]/\PP(S=0), \label{eq:thing-to-bound}
\end{align}
where \eqref{eq:thing-to-bound} holds since
\begin{align*}
  \PP(R_a(Y(a),\bm V) \leq  r)  | S=0 ) - (1-\alpha) &=
  \E[m_a(r,\bm V) - (1-\alpha)| S=0] \\
  &=
  \E[\chi_a(r,O ; \eta_a(r))]/\PP(S=0),
\end{align*}
for any $r$.

Thus, demonstrating (\ref{eq:cov-rate}) amounts to showing 
\begin{equation}
     \E[\chi_a(\bm O,\hatra;\eta)] = O_\PP(1/\sqrt n + R_n).
\end{equation}

We consider the following decomposition for $\E[\chi_a(\bm O,\hatra; \eta)]$. For brevity, we omit the observational arguments and define $\E[\chi_a(\hatra, \eta)]  := \E[\chi_a(\bm O,\hatra; \eta)]$, noting 
\begin{align*}
    \E[\chi_a(\hatra,\eta)] 
    &= 
    \E(\chi_a(\hatra,\hat \eta) - \chi_a(\hatra, \eta)) \\
&- (\PP_n-\E)[\chi_a(\hatra,\hat \eta)] \\
&+ \PP_n(\chi_a(\hatra,\hat \eta))
\end{align*}
Above,  the third term is zero by construction.
The second term  is $O_\PP(1/\sqrt n)$ if either (i) $\hat \psi_a(r_{a,\alpha}, \hat \eta)$ lies in a Donsker class \citep{van2000asymptotic}, or (ii) if $\hat \eta$ is obtained from a separate sample \citep{kennedy2020efficient}. Since we employ the cross-fitting procedure suggested by \cite{kallus2024localized}, condition (ii) holds regardless of whether all relevant nuisance functions fall into a Donsker class, implying the second term above is $O_\PP(1/\sqrt n)$. We note that modest assumptions on the nuisance functions $\hat \eta$ employed in related work \citep{liu2024multi} additionally ensure this rate of convergence without the need for cross fitting, but these assumptions are not strictly necessary given our use of cross-fitting.

We turn our focus to the first term above, $\E(\chi_a(\hatra,\hat \eta) - \chi_a(\hatra, \eta))$. 

Our strategy for bounding this first term closely follows that of \cite{zeng2025efficient}. Notice for any generic $r$, we have
\begin{align*}
 &  \E(\chi_a(r,\hat \eta) - \chi_a(r, \eta)) \\
 &=
   \E[(1-S)(\hat m_a(r,\bm V) - m_a(r,\bm V)] \\
   &+
   \E\left[ \frac{S(1-\hat \kappa(\bm V))}{\hat \kappa(\bm V)}(\tilde m_a(r,\bm V) -  \hat m_a(r,\bm V)) \right] \\
   &+
\E\left[ \frac{\mathbb{I}(A=a)S(1-\hat \kappa(\bm V))}{\hat \kappa(\bm V) \hat g_a(\bX)}( q_a(r,\bX) - \hat q_a(r,\bm X)) \right],    
\end{align*}
where $\tilde m_a(r,\bm V) = \E[\hat q_a(r,\bX)]$. We can remove dependence on $\tilde m_a(r,\bm V)$ by noting the second term can be rewritten as
\[
 \E\left[ \frac{S(1-\hat \kappa(\bm V))}{\hat \kappa(\bm V)}(\tilde m_a(r,\bm V) - m_a(r,\bm V)) \right]
 - 
  \E\left[ \frac{S(1-\hat \kappa(\bm V))}{\hat \kappa(\bm V)}(\hat m_a(r,\bm V) - m_a(r,\bm V)) \right],
\]
where we can then leverage the fact that
\[
 \E\left[ \frac{S(1-\hat \kappa(\bm V))}{\hat \kappa(\bm V)}(\tilde m_a(r,\bm V) - m_a(r,\bm V)) \right] = 
  \E\left[ \frac{S(1-\hat \kappa(\bm V))}{\hat \kappa(\bm V)}(\hat q_a(r,\bm X) - q_a(r,\bm X)) \right].
\]
Given this form for the second term, after re-arranging we can rewrite $\E(\chi_a(r,\hat \eta) - \chi_a(r, \eta))$ as
\begin{align*}
 &  \E(\chi_a(r,\hat \eta) - \chi_a(r, \eta)) \\
 &=
  \E\left[\frac{(1-\hat \kappa(\bm V))(\hat q_a(r,\bm X) - q_a(\bm X)) }{\hat \kappa(\bm V)}\left\{S - \frac{\mathbb{I}(A=a)S}{\hat g_a(\bX)} \right\} \right] \\
  &+ 
  \E\left[ \left\{
  (1-S) - \frac{S(1-\hat \kappa(\bm V))}{\hat \kappa(\bm V)}
  \right\} (\hat m_a(\bm V) - m_a(\bm V))
  \right] \\
  &= \text{I} + \text{II}.    
\end{align*}
Now, term I above can be bounded by noting
\begin{align*}
    \text{I} &=\E\left[\frac{(1-\hat \kappa(\bm V))(\hat q_a(r,\bm X) - q_a(r,\bm X)) }{\hat \kappa(\bm V)}\left\{S - \frac{\mathbb{I}(A=a)S}{\hat g_a(\bX)} \right\} \right] \\
 &=
 \E\left[\frac{\PP(S=1|\bX)(1-\hat \kappa(\bm V))(\hat q_a(r,\bm X) - q_a(r,\bm X)) }{\hat \kappa(\bm V)\hat g_a(\bX)}\left\{ \hat g_a(\bX) - g_a(\bX)\right\} \right] \\
 &\leq 
 \frac{1}{\varepsilon'} \E[(\hat q_a(r,\bm X) - q_a(r,\bm X)) \cdot (\hat g_a(\bX) - g_a(\bX)) ] \\
& \leq
\frac{1}{\varepsilon'} ||\hat q_a(r,\bm X) - q_a(r,\bm X) || \cdot ||  \hat g_a(\bX) - g_a(\bX)||,
\end{align*}
where $\varepsilon' >0$.
Above, the third line holds by positivity conditions outlined at the beginning of the proof, while the fourth line holds by the Cauchy-Schwarz inequality. 

Through similar logic, we  can bound the second term by noting
\begin{align*}
 \text{II} &= 
 \E\left[ \left\{
  (1-S) - \frac{S(1-\hat \kappa(\bm V))}{\hat \kappa(\bm V)}
  \right\} (\hat m_a(\bm V) - m_a(\bm V))
  \right] \\
  &=
  \E \left[ 
  \frac{(\hat \kappa(\bm V) - \kappa(\bm V))(\hat m_a(r,\bm V) - m_a(r, \bm V))}{\hat \kappa(\bm V)}
  \right]
   \\
  &\leq
  \frac{1}{\varepsilon'} || \hat \kappa - \kappa || \cdot || \hat m_a(r) - m_a(r) ||
\end{align*}
Notice I and II imply that 
\begin{align*}
 \E(\chi_a(\hatra,\hat \eta) - \chi_a(\hatra, \eta)) & \leq \sup_r  \frac{1}{\varepsilon'} ||\hat q_a(r) - q_a(r) || \cdot ||  \hat g_a - g_a|| + \sup_r 
   || \hat \kappa - \kappa || \cdot || \hat m_a(r) - m_a(r) || \\
 &= O_\PP\left(\sup_r  ||\hat q_a(r) - q_a(r) || \cdot ||  \hat g_a - g_a|| + \sup_r 
   || \hat \kappa - \kappa || \cdot || \hat m_a(r) - m_a(r) ||\right)
\end{align*}
where we use the fact that by construction $||\hat q_a(\hat r_{a,\alpha}) - q_a(\hat r_{a,\alpha}) || \leq \sup_r||\hat q_a(r) - q_a(r)||$, analogously holding for $\hat m_a(r)$
Recalling $\PP(Y(a) \in \hat C_a(\bm V) | S=0) -  1-\alpha = \E[\chi_a(\bm O,\hatra;\eta)]/\PP(S=0)$ and the decomposition of $\E[\chi_a(\bm O,\hatra\;\eta)]$ yields the desired result.

\clearpage

\section{Additional Experiments Details}
\label{sec:exp-details}

\subsection{Methods Implementation}

We briefly provide additional information on the implementation of the naive DML estimator which ignores runtime confounding, and the weighted estimator explored throughout our numerical experiments. Details on specific training parameters are provided later in the section.

In implementing the naive DML estimator, we follow Algorithm \ref{alg:alg-debiased}, enforcing $\bm X=\bm V$. In the setting where one forces $\bX = \bm V$ by ignoring runtime confounding, the EIC reduces to
\begin{align}
\chi_a(r_{a,\alpha},  \bm O\  ; \eta_a(r_{a,\alpha})) 
=
(1-S)(q_a(r_{a,\alpha},\bm V) - (1-\alpha))  
+  w_a(\bm O) \{\II(R_a(Y,\bm V)\leq r_{a,\alpha}) - q_a(r_{a,\alpha},\bm V) \}, \nonumber
\end{align}
where $\tilde w_a(\bm O) = \frac{A S (1-\kappa(\bm V))}{\tilde g_a(\bm V)\kappa(\bm V)}$,
where $\tilde g_a(\bm V) := \PP(A=a|\bm V, S=1)$. Intuitively, with this restriction we effectively have $m_a(r,\bm V) = q_a(r,\bm V)$, canceling out middle term in the original EIC. 

The weighted estimator is obtained through a split conformal prediction procedure which solves the estimating equation implied by Equation \ref{eq:thm1-ipw} on the calibration fold of source observations.

\subsection{Numerical Experiments}

In this section, we provide full details on the procedures used to generate data in our numerical and semi-synthetic experiments in \ref{sec:simulation}. 

\subsubsection{Data Generation}

Our numerical experiments extend the setup considered in \cite{coston2020counterfactual}. Letting $\bm V = (V_1,\ldots, V_{p_V})$ and  $\bm U = (U_1,\ldots, U_{p_U})$ 
\begin{align*}
    V_k &\sim \mathcal{N}(0,1),   \quad 1 \leq k \leq p_V \\
    U_k &\sim \mathcal{N}(0, 1), \quad  1 \leq k \leq p_U \\
     Y(a) &= \mu(\bm V,\bm U) + \epsilon(\bm V,\bm U), \quad
     \mu(\bm V,\bm U) = \frac{k_V}{k_V + k_U} \left( \sum_{k=1}^{k_V} V_k + 2\sum_{k=1}^{k_U} U_i \right) 
    \\
    A &\sim \text{Bernoulli}(\pi(\bm V,\bm U)), \quad \quad g(\bm V,\bm U) = \text{expit} \left( \frac{1}{\sqrt{k_V + k_U}}\left( \sum_{i=1}^{k_V} V_i - 2\sum_{i=1}^{k_U} U_i \right) \right) \\
    S &\sim \text{Bernoulli}(\kappa (\bm V)), \quad \quad \kappa(\bm V) = \text{expit}\left(b - \frac{1}{\sqrt{ k_V}}\sum_{k=1}^{k_V} V_k \right)
\end{align*}
where $\text{expit}(x) = \exp(x)/(1+\exp(x))$, $\epsilon(\bm V, \bm U) \sim N(0, \sqrt{|\mu(\bm V, \bm U) |})$
$\bm V = (V_1,\ldots,V_{k_V})$, $\bm U = (U_1,\ldots,U_{k_U})$, $k_V \leq p_V$ and $k_U \leq p_U$. We choose $b$ in $\kappa(\bm V)$ to ensure $\E[\kappa(\bm V)] = \PP(S=1) = 0.9$, achieving this numerically by simulating 1 million values of $\bm V$ outside of our main simulation.

Notably, source population membership is influenced by $\bm V$,  generating covariate shift between the source and target populations. $A$ and $Y(a)$ are both influenced by $\bm V$ and $\bm U$. To induce runtime confounding, we treat $\bm U$ as unobserved in the target population ($S=0$). 

We set $p_V = p_U = 15$, $k_V = 5$ and vary $k_U \in \{5,10,15\}$. This setup induces sparsity in the outcome, treatment and population models, while allowing us to investigate the impact of increasingly severe instances of runtime confounding.

We allow for covariate shift between $S=0$ and $S=1$ units by simulating $S$ as a function of $\bm V$, extending the setup considered in \cite{coston2020counterfactual}.

\subsubsection{Training Details}

Constructing prediction intervals among the three approaches considered requires estimation of $g_a, \kappa, q_a$ and $m_a$. We additionally require estimation of $\mu_a$ and $\eta_a$ when using absolute residual conformity scores, and estimation of $Q_{a,\alpha}$ when using quantile conformity scores.

We fit all of $g_a, \kappa, q_a,$ $m_a,$  with a stacked ensemble of random forests and Lasso models. 
We fit these ensemble learners with the \texttt{SuperLearner} package in R. Random forests are fit with the \texttt{ranger} package, using default hyperparameters specified by the \texttt{SL.ranger} SuperLearner library. 
Lasso models are fit with the \texttt{glmnet} package, similarly choosing default values specified by \texttt{SL.glmnet}.  Both estimated treatment and source probability are trimmed to be within the interval $(0.025,0.975)$ to avoid instability induced by large inverse propensity weights.

When using absolute residual conformity scores, we use the two-stage procedure proposed by \cite{coston2020counterfactual} and described in Section \ref{sec:conformal-background}. In this setting, $\mu_a$ and $\eta_a$ are fit with this same stacked ensemble with \texttt{SuperLearner}. When using methods which ignore runtime confounding, effectively $\mu_a=\eta_a$, meaning we only fit  $\mu_a$.

When using quantile conformity scores, we fit $Q_{a,\alpha}$ with weighted quantile forests, using the weights we propose in Proposition \ref{prop:wgtd-quant}. We use weights of the form $\hat w_a(\bm O)$, recalling $\hat w_a(\bm O)$ is a function of $\hat g_a$ and $\hat \kappa$ fit according to the procedure above, and use the \texttt{ranger} package to implement the corresponding weighted quantle regression, specifying the same parameters as above. When using methods which ignore runtime confounding, we perform unweighted quantile regression.

\subsection{Data Application}
\label{ap:data-app}

\subsubsection{Data Generation Details}

We emulate the data generating procedure employed by \cite{lei2021conformal}, additionally enforcing runtime confounding. We describe the procedure here, emphasizing that the data generation closely follows the procedure described in \cite{lei2021conformal}. Following \cite{lei2021conformal}, we split the ACIC data into two folds, $Z_1$ and $Z_2$, where $|Z_1|=2079$ and $|Z_2|=8312$. 

To investigate varying degrees of runtime confounding, we consider three splits of $\bX = (\bm V, \bm U)$:
\begin{enumerate}
    \item \textbf{Severe}:  $\bm V = (\texttt{X3},\texttt{X4},\texttt{X5},\texttt{XC})$ and $\bm U = (\texttt{X1},\texttt{X2},\texttt{X5},\texttt{C1},\texttt{C3},\texttt{S3})$
    \item \textbf{Moderate}: $\bm V = (\texttt{X1}, \texttt{X2}, \texttt{X3}, \texttt{X4}, \texttt{X5}, \texttt{XC})$, $\bm U = (\texttt{C1}, \texttt{C2}, \texttt{C3}, \texttt{S3})$
    \item \textbf{Mild}: $\bm V = (\texttt{X1}, \texttt{X2}, \texttt{X3}, \texttt{X4}, \texttt{X5}, \texttt{XC}, \texttt{S3}, \texttt{C1})$, $\bm U = (\texttt{C2},\texttt{C3})$
\end{enumerate}

On $Z_1$ and for each runtime confounding scenario we consider, we 
\begin{itemize}
    \item Fit $\hat m_0(\bX) = \hat \E[Y(0) | \bX]$ with the \texttt{randomForest} package. 
    \item We fit $\hat g(\bX)=\hat \PP(A=1|\bX)$ through the \texttt{randomForest} package, truncating $\hat g(\bX)$ to fall within 0.1 and 0.9
    \item We fit the 25\% and 75\% conditional quantile functions for $Y(0)$ and $Y(1)$ with the \texttt{grf} package, and let  $\hat r_0(\bX)$ and $\hat r_1(\bX)$ denote the corresponding interquartile ranges 
    \item We regress \texttt{Z} on $\bm V$ with the \texttt{randomForest} package, obtaining predictions $\hat \kappa(\bm V)$. Treating the predicted probabilities as $\tilde \kappa(\bm V)$, we produce $\kappa(\bm V)$ by choosing $b$ such that $\E[\text{expit}(b - \text{logit}(\tilde \kappa(\bm V))] = 0.9$.
\end{itemize}
We then let
\[
Y_i(0) = \hat m_0(\bX_i)  + 0.5\hat r_0(\bX_i)\varepsilon_{i0}, \ \ \  Y_i(1) = \hat m_1(\bX_i) + \tau(\bX_i) + 0.5\hat r_1(\bX_i)\varepsilon_{i1},
\]
where $\varepsilon_{ia}$ are iid $N(0,1)$ for $a=0,1$ and $\tau(\bX)$ is the CATE function defined in equation (1) of \cite{carvalho2019assessing}. We then simulate data according to 
\begin{align*}
    \bX &\sim F_{Z_2} \\
    A|\bX &\sim \text{Bernoulli}(\hat g(\bX)) \\
    Y_i(0) &= \hat m_0(\bX_i)  + 0.5\hat r_0(\bX_i)\varepsilon_{i0} \\  Y_i(1) &= \hat m_1(\bX_i) + \tau(\bX_i) + 0.5\hat r_1(\bX_i)\varepsilon_{i1}, \\
    S &\sim \text{Bernoulli}(\hat \kappa(\bm V)),
\end{align*}
where $F_{Z_2}$ is the empirical distribution of covariates in the held out split of data $Z_2$. Note that we enforce a runtime confounding scenario by simulating source population membership through $S$ as a function of $\bm V$, extending the setup considered in \cite{lei2021conformal}. 

In implementing our considered methods, nuisance functions, we train nuisance functions using the same learners considered for our numerical simulations. 

\clearpage

\section{Additional Numerical Experiment Results}
\label{sec:extra-exp-results}

\subsection{Results Stratified by Treatment Level}
\begin{figure}[h!]
    \centering
    \includegraphics[scale=0.55]{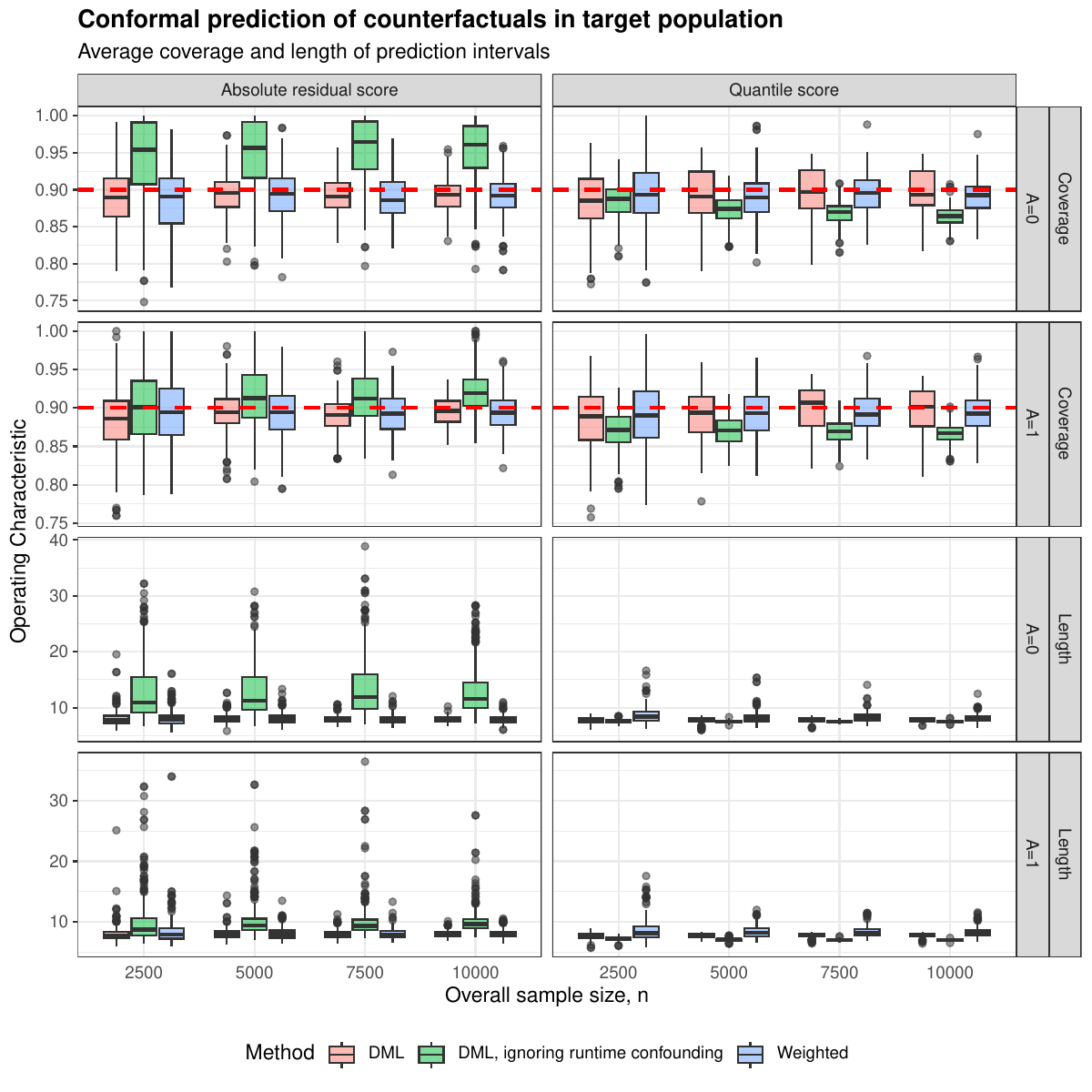}
    \caption{Performance of proposed methods stratified by counterfactual outcome.}
    \label{fig:sim-res-byA}
\end{figure}

Our main results pool coverage rates and average interval lengths for both $Y(1)$ and $Y(0)$ in the target population. Figure \ref{fig:sim-res-byA} reports coverage rates and interval lengths separately for $Y(1)$ and $Y(0)$. Qualitative results are similar. Average interval lengths are typically larger for $Y(1)$.

\clearpage

\subsection{Varying the Degree of Runtime Confounding}
We report results stratified by treatment level $a$ when varying the degree of true runtime confounders, controlled by $k_V$ in Section \ref{sec:exp-details}. Results remain qualitatively similar to our baseline scenario where $k_V=10$.

\begin{figure}[h!]
    \centering
    \includegraphics[width=0.6\linewidth]{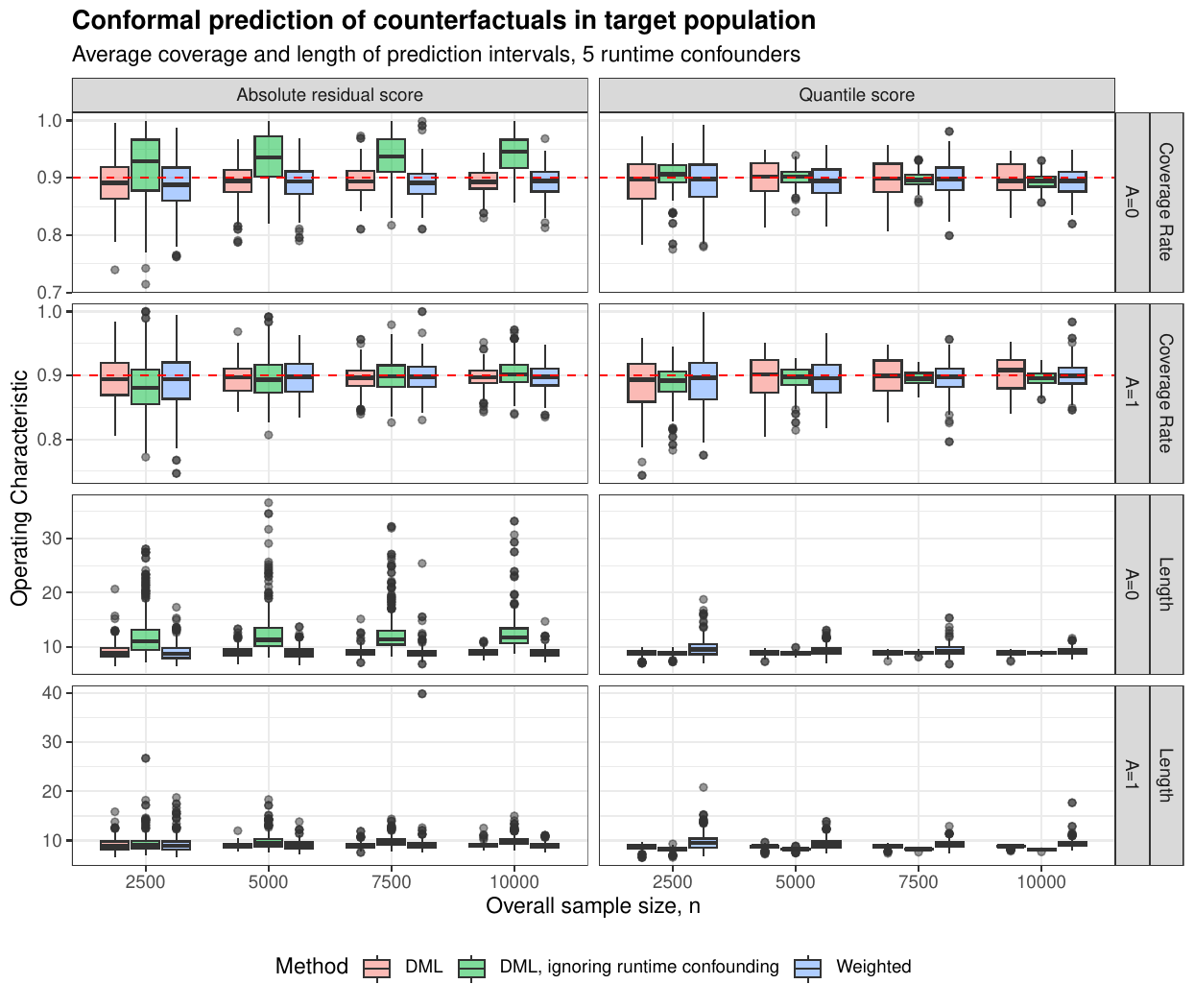}
    \caption{Performance of proposed methods, varying $n$ and fixing the number of runtime confounders at 5.}
    \label{fig:rc5}
\end{figure}

\begin{figure}[h!]
    \centering
    \includegraphics[width=0.6\linewidth]{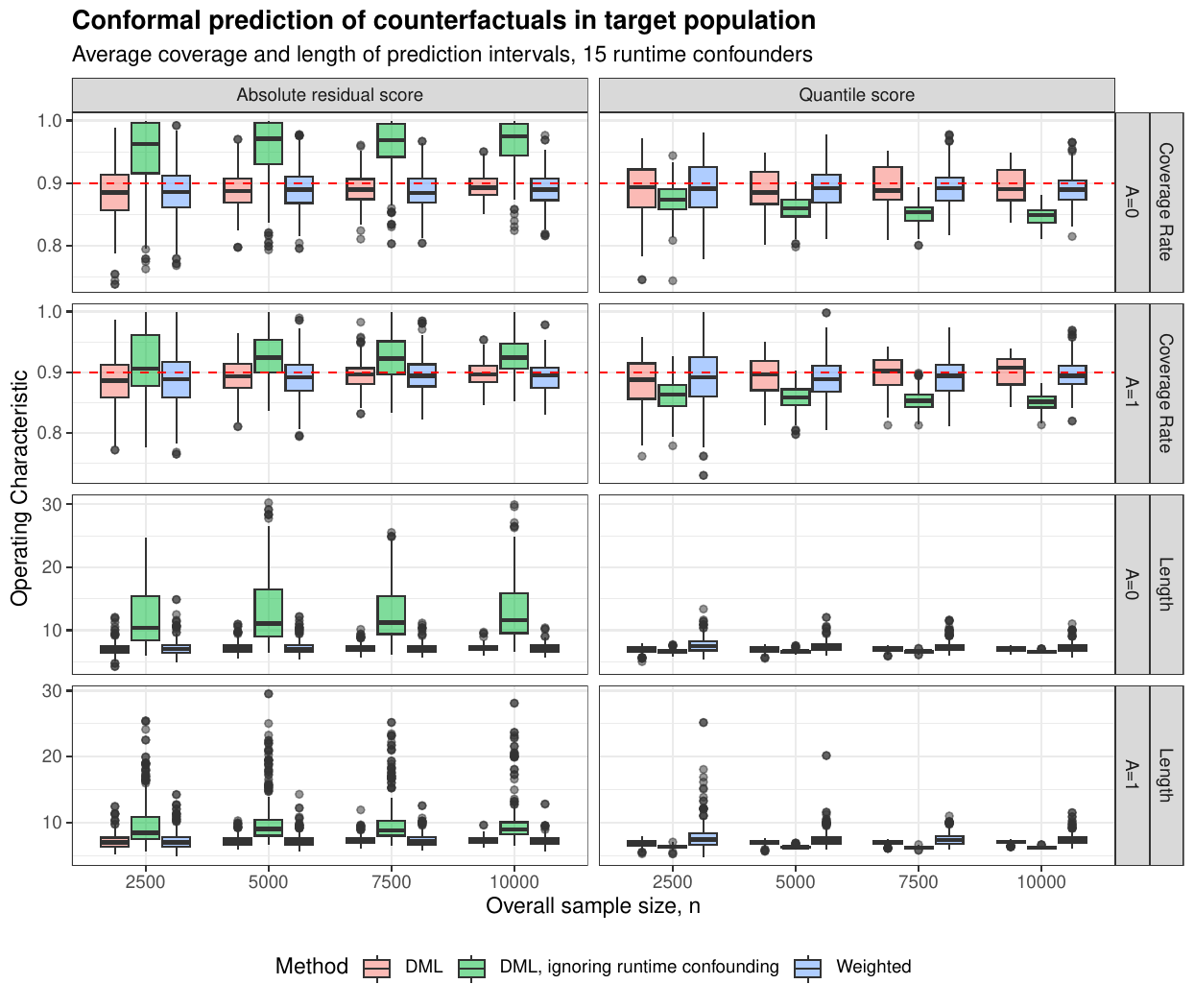}
    \caption{Performance of proposed methods, varying $n$ and fixing the number of runtime confounders at 15.}
    \label{fig:rc15}
\end{figure}


\subsection{Varying the Share of Target Population Data}

Fixing $n=5000$ and the number of runtime confounders at $10$, we vary the share of source data $\PP(S=1)$, finding similar results across all shares considered.

\begin{figure}
    \centering
    \includegraphics[width=0.8\linewidth]{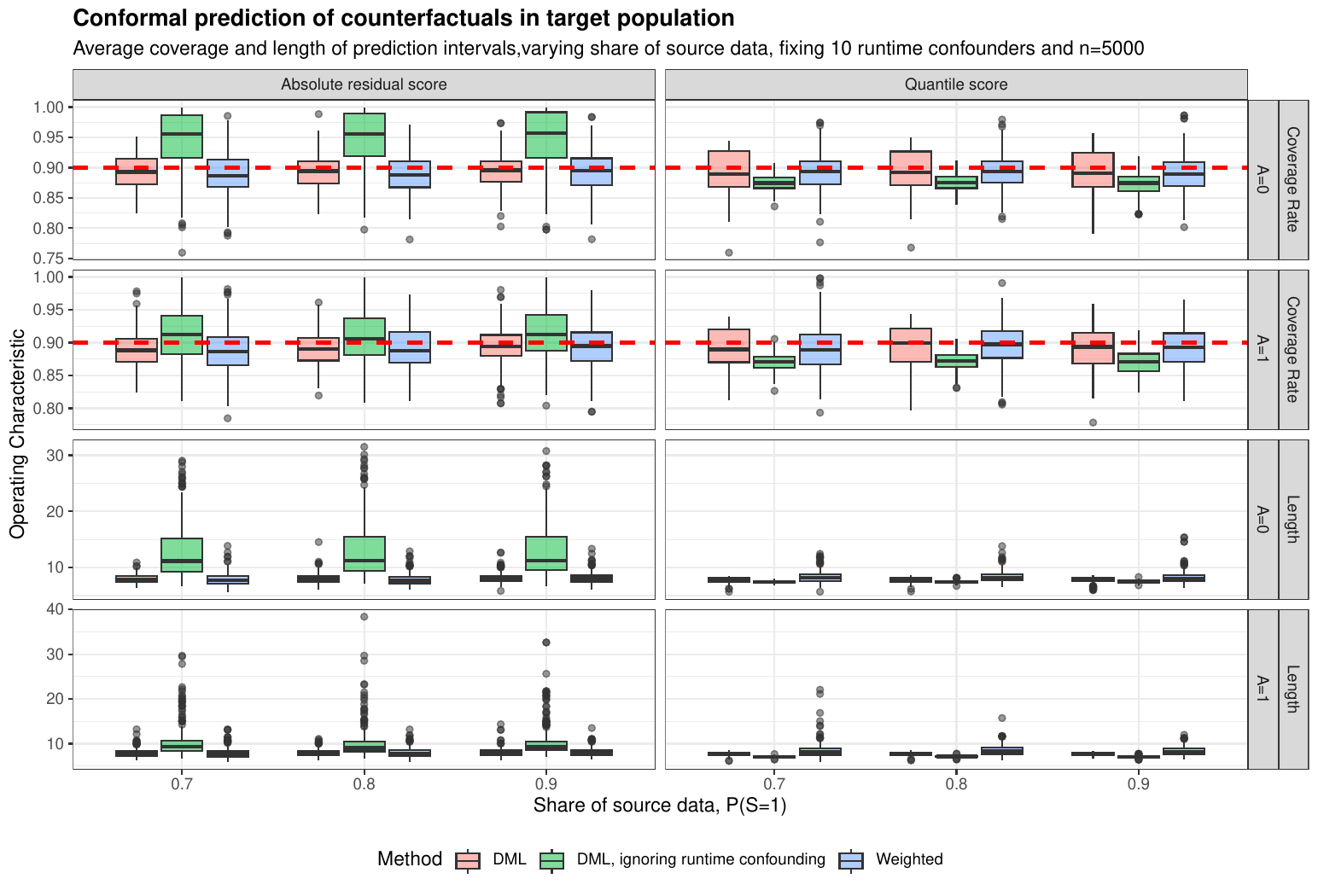}
    \caption{Performance of proposed methods when varying $\PP(S=1)$, fixing $n=5000$ and the number of runtime confounders at 10.}
    \label{fig:placeholder}
\end{figure}

\clearpage

\section{Additional Semi-Synthetic Experiment Results}
\label{ap:extra-data-app-results}

\subsection{Baseline Results Stratified by Treatment Level}

\begin{figure}[h!]
    \centering
    \includegraphics[width=0.6\linewidth]{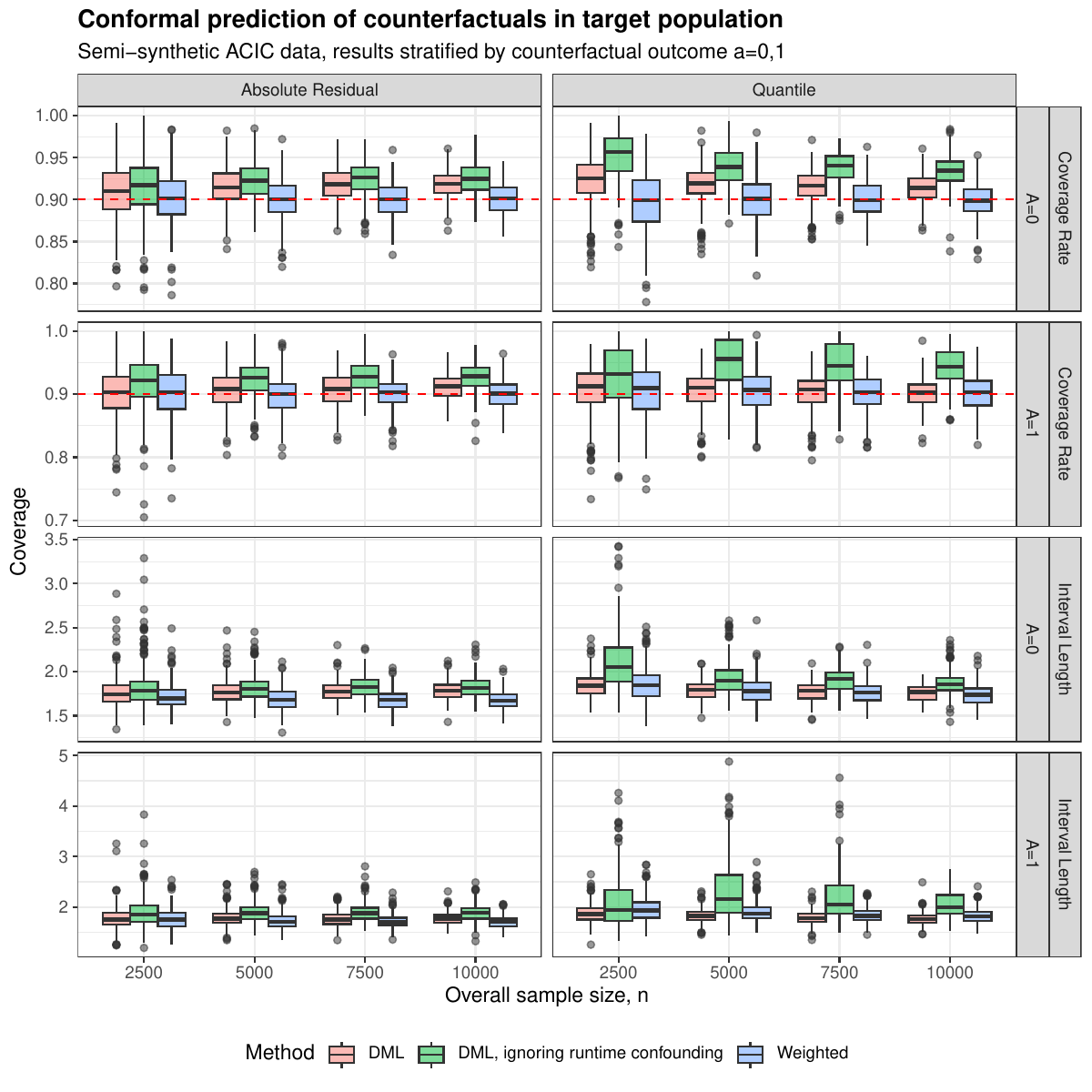}
    \caption{Performance of proposed methods on semi-synthetic ACIC data, varying $n$ under the baseline moderate runtime confounding scenario.}
    \label{fig:placeholder}
\end{figure}

\subsection{Varying Degree of Runtime Confounding}
In this section, we report the results obtained by repeating our semi-synthetic ACIC data exercise when the set of variables included in $\bm V$ varies as outlined in Appendix \ref{ap:data-app}. Intuitively, the fewer covariates available in $\bm V$, the greater the degree of runtime confounding. We see that the naive DML approach deteriorates in the severe runtime confounding scenario, often producing excessively wide intervals relative to the weighted and DML approaches.

\begin{figure}[h!]
    \centering
    \includegraphics[width=0.6\linewidth]{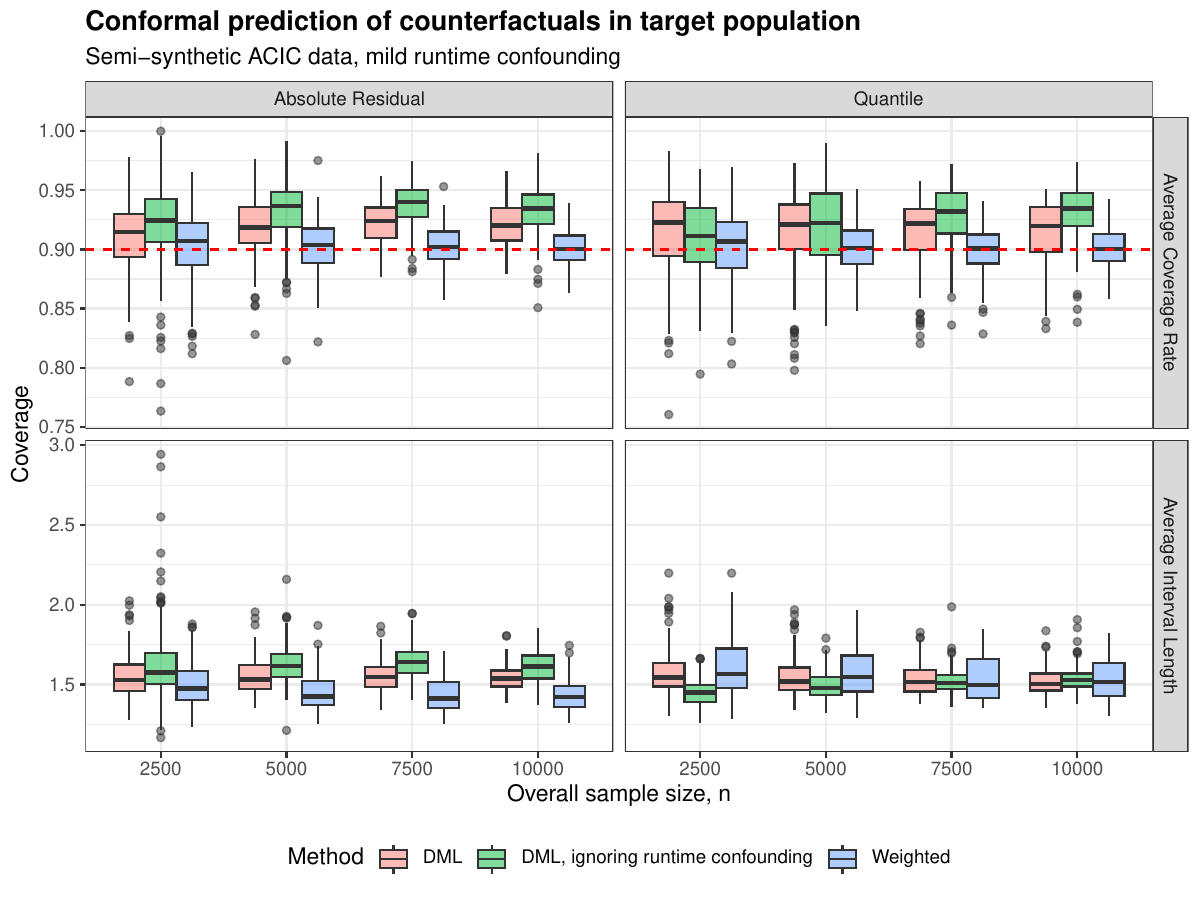}
    \caption{Performance of proposed methods on semi-synthetic ACIC data under the mild runtime confounding scenario.}
    \label{fig:placeholder}
\end{figure}

\begin{figure}[h!]
    \centering
    \includegraphics[width=0.6\linewidth]{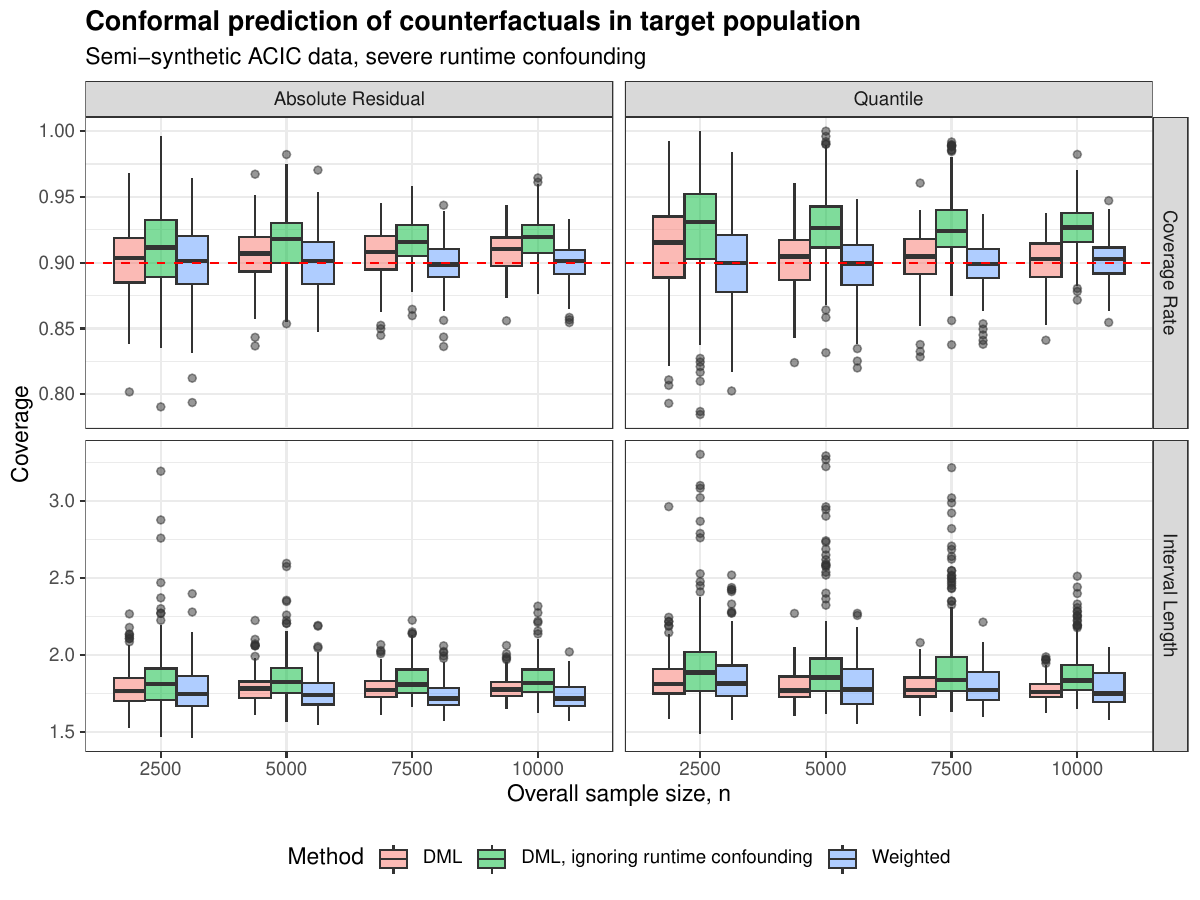}
    \caption{Performance of proposed methods on semi-synthetic ACIC data under the severe runtime confounding scenario.}
    \label{fig:placeholder}
\end{figure}

\clearpage 

\section{Intervals for Individual Treatment Effects}
\label{ap:ite-methods}

While we fixate interest on the setting where $A$ is a categorical random variable and interest lies in the construction of intervals for $Y(a)|S=0$ for generic $a$, a large set of work has covered the setting where $A \in \{0,1\}$ and interest lies in constructing intervals for  target population individual treatment effects $Y(1)-Y(0)|S=0$.

Following \cite{lei2021conformal}, a straightforward approach to construct prediction intervals $\hat C_\text{ITE}(\bm V)$ targeting the coverage result
$$
\PP(Y(1)-Y(0) \in \hat C_\text{ITE}(\bm V) | S=0)=1-\alpha
$$
is to
\begin{enumerate}
    \item Construct $1-\alpha/2$ level intervals $\hat C_1(\bm V) = (\hat C_1^\text{L}(\bm V), \hat C_1^\text{U}(\bm V))$ and $\hat C_0(\bm V) = (\hat C_0^\text{L}(\bm V), \hat C_0^\text{U}(\bm V))$ for $Y(1)$ and $Y(0)$, respectively, using Algorithm \ref{alg:alg-debiased}
    \item Construct intervals of the form $\hat C_\text{ITE}(\bm V) = (\hat C_1^\text{L}(\bm V)-\hat C_0^\text{U}(\bm V), \hat C_1^\text{U}(\bm V)-\hat C_0^\text{L}(\bm V))$
\end{enumerate}
Although easy to implement, the above approach will tend to produce excessively wide intervals. Alternatively, one can construct nested intervals as outlined in \cite{lei2021conformal} and later extended to handle target-source covariate shift in a surrogate outcome setting by \cite{gao2025bridging}. Although not the focus of this work, we briefly discuss the high-level procedure one can follow: 
\begin{enumerate}
    \item Within the source population, construct intervals $\hat C(\bm X)$ which aim to satisfy
    $$
    \PP(Y(1)-Y(0) \in \hat C(\bX) | S=1)
    $$
    To do this, suppose  $C_a(\bX)$ satisfies $\PP(Y(a) \in C_a(\bX) | S=1,A=1-a)$. Since $A$ is observed for all units in the source population, one can construct ITE intervals in the source population of the form
    $$
  C(\bX) =
    \begin{cases}
    Y- C_0(\bX), &  A\cdot S=1, \\
    C_1(\bX)-Y, & (1-A)\cdot S =1
    \end{cases}
    $$
    The component intervals $C_a(\bX)$  can be constructed using the doubly-robust procedure proposed in \cite{yang2024doubly}, where all of $\bX$ can be used since $\bX$ is available for all members of the source population.
    \item Define a conformity score $R_C(C,\bm V)$ with respect to the individual-level intervals $\hat C(\bX_i)$ in the source population. \cite{gao2025bridging} provide recommendations for choices of scores, where here we restrict the scores to incorporate only $\bm V$ since $\bm U$ is unobserved in the target population
    \item Target the $1-\gamma$ quantile of $R_C$ in the target population, denoted $r_\gamma$ which satisfies
    $$
    \PP(R_C(C,\bm V) \leq r_\gamma | S=0) = 1-\gamma,
    $$
    noting under the earlier independence assumptions we will have $r_\gamma$ additionally satisfies
    $$
    \E[\PP(R_C(C,\bm V) \leq r_\gamma | S=1,\bm V) | S=0] = 1-\gamma.
    $$
\end{enumerate}
Given the above identifying functional, one can construct doubly-robust estimators of $r_\gamma$ using the approach outlined in \cite{gao2025bridging}, forming intervals of the form $C_\text{ITE}(\bm V) = \{c: R_C(c,\bm V) \leq \hat r_\gamma \}$, 
who established the resulting intervals asymptotically satisfy 
$$
\PP(Y(1) - Y(0) \in C_\text{ITE}(\bm V) | S=0) \geq 1 - (\alpha+\gamma).
$$
under standard regularity conditions.
While the above procedure will yield intervals with the desired properties, we devote a formal implementation and study of the resulting intervals to future work.

\clearpage

\section{Discussion of the Independence Assumption \ref{as:source-exch}}
\label{sec:app-ind-discussion}

\begin{figure}[h!]
\centering
\subfigure[]{
    \resizebox{0.27\linewidth}{!}{%
    \begin{tikzpicture}[node distance=0.8cm, every node/.style={transform shape}]
        \node[circle,draw,minimum size=0.7cm] (V) at (-0.5,1.5) {$\bm V$};
        \node[circle,draw,minimum size=0.7cm,fill=gray!30] (U) at (-0.5,-1.5) {$\bm U$};
        \node[circle,draw,minimum size=0.7cm] (S) at (-2,0){$S$};
        \node[circle,draw,minimum size=0.7cm] (A) at (2,0) {$A$};
        \node[circle,draw,minimum size=0.7cm] (Y) at (4,0) {$Y$};
        \draw[->, thick, >=stealth] (V) -- (A);
        \draw[->, thick, >=stealth] (S) -- (V);
        \draw[->, thick, >=stealth] (V) -- (Y);
        \draw[->, thick, >=stealth] (U) -- (A);
        \draw[->, thick, >=stealth] (U) -- (Y);
        \draw[->, thick, >=stealth] (S) -- (A);
        \draw[->, thick, >=stealth] (A) -- (Y);
        \draw[->, thick, >=stealth] (S) -- (U);
    \end{tikzpicture}}}
    \hspace{8em}
\subfigure[]{
    \resizebox{0.27\linewidth}{!}{
        \begin{tikzpicture}[node distance=0.8cm, every node/.style={transform shape}]
        \node[circle,draw,minimum size=0.7cm] (V) at (-2,1.5) {$\bm V$};
        \node[circle,draw,minimum size=0.7cm,fill=gray!30] (U) at (-2,-1.5) {$\bm U$};
        \node[circle,draw,minimum size=0.7cm] (S) at (-0.5,0){$S$};
        \node[circle,draw,minimum size=0.7cm] (A) at (2,0) {$A$};
        \node[circle,draw,minimum size=0.7cm] (Y) at (4,0) {$Y$};
        \draw[->, thick, >=stealth] (V) -- (A);
        \draw[->, thick, >=stealth] (V) -- (S);
        \draw[->, thick, >=stealth] (V) -- (Y);
        \draw[->, thick, >=stealth] (U) -- (Y);
        \draw[->, thick, >=stealth] (U) -- (A);
        \draw[->, thick, >=stealth] (S) -- (A);
        \draw[->, thick, >=stealth] (A) -- (Y);
        \draw[->, thick, >=stealth] (S.north)  to [out=50] (Y.north);
    \end{tikzpicture}}}
    \caption{Two possible directed acyclic graphs consistent where Assumption \ref{as:assumption-unconfoundedness} is satisfied but \ref{as:source-exch} is violated.}
    \label{fig:dag_grid-vio}
    \vspace{-2em}
\end{figure}
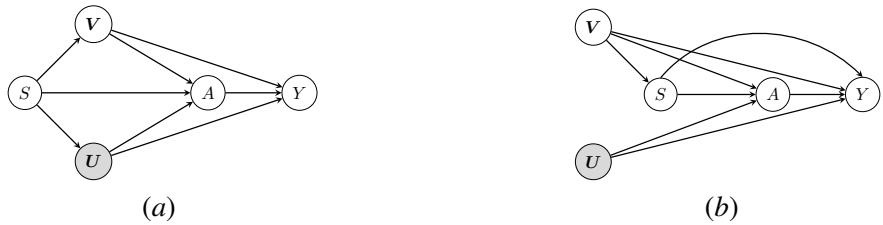

As discussed in Section \ref{sec:prob-setting}, it can be difficult to assess the plausibility of Assumption \ref{as:source-exch} in multi-source settings. In this Section, we briefly discuss recommendations for assessing the plausibility of this Assumption.
\\ \\
We begin by noting Assumption \ref{as:source-exch} is implied by the following two alternative Assumptions:
\begin{intassumption}
    \label{as:ySX}
   $Y(a) \indep S | \bX$  
\end{intassumption}
\begin{intassumption}
    \label{as:alt-covUS}
    $\bm U \indep S | \bm V$
\end{intassumption}
To see this, assume for simplicity that the data are discrete and note that for any $y,v,s$ 
\begin{align*} \mathbb{P}(Y(a) = y) | \mathbf{V}=v,S=s) &= \sum_u \mathbb{P}(Y(a) = y) \big| \mathbf{V}=v,\mathbf{U}=u,S=s)\mathbb{P}(\mathbf{U}=u|V=v,S=s) \\ &= \sum_u \mathbb{P}(Y(a) = y) \big| \mathbf{V}=v,\mathbf{U}=u)\mathbb{P}(\mathbf{U}=u|V=v), 
\end{align*} 
where the last display does not depend on $s$, implying $Y(a) \indep S | \bm V$, which is exactly Assumption \ref{as:source-exch}. Assumption \ref{as:ySX} can be viewed as a weaker version of Assumption \ref{as:source-exch} that conditions on the full set of covariate information, which in tandem with Assumption \ref{as:assumption-unconfoundedness} implies that the set of covariates $\mathbf{X}$ that are sufficient to control for treatment-outcome confounding in the source population are additionally sufficient to render $Y(a)$ independent from $S$. Relatedly, Assumption \ref{as:alt-covUS} implies there is no covariate shift in $\mathbf{U}$ across populations conditional on the always-observed $\mathbf{V}$, which may be plausible in settings where the source and target sites do not enroll.
\\ \\
While we believe it often easiest to assess the plausibility of Assumption \ref{as:source-exch} through the plausibility of both , since we rely on their implied condition $Y(a) \indep S | \bm V$ for identification, we invoke this condition directly in the manuscript. In light of this alternative framing, we discuss examples in which we expect Assumption \ref{as:source-exch} to hold below, and provide example DAGs where Assumption \ref{as:source-exch} is violated in Figure \ref{fig:dag_grid-vio}.

\subsection*{Example Scenarios where Assumption \ref{as:source-exch} will be Plausible}

To develop intuition for determining the plausibility of Assumption \ref{as:source-exch}, consider a runtime confounding setting involving the treatment of acute ischemic stroke. Interest lies in forming counterfactual prediction intervals for the impact of different treatments $A$, (e.g. thrombectomy) on hospital length of stay 
among individuals receiving care from a target population hospital, using data from a separate hospital corresponding to the source population.
\\ \\
Suppose $\bm V$ collects baseline demographic characteristics and readily obtainable information including blood pressure, age, and NIH stroke scale. Further suppose $\bm U$ contains additional information which informs treatment decisions in the source population---such as cerebral blood flow---but is more resource-intensive to collect and in turn not readily available in the target population hospital. Assumption \ref{as:ySX} will be plausible if $\bm V$ and $\bm U$  explain away hospital-specific effects on length of stay, and Assumption \ref{as:alt-covUS} will be plausible if the target and source population hospitals enroll similar patient populations at baseline. Recall that these two Assumptions in turn imply the desired condition $Y(a) \indep S | \bm V$.
\\ \\
Alternatively, Assumption \ref{as:alt-covUS} may be less plausible if the target and source hospitals enroll patients with notably different baseline characteristics, and Assumption \ref{as:ySX} may be less plausible if features unmeasured in both sites but tied to hospital quality---such as staff size---meaningfully influence length of stay.

\end{document}